\pdfoutput=1
\documentclass[10pt,twocolumn,letterpaper]{article}

\usepackage[pagenumbers]{cvpr}

\usepackage{microtype}

\renewcommand{\paragraph}[1]{\vspace{.5em}\noindent\textbf{#1.}}

\usepackage{colortbl}
\usepackage{stfloats}
\definecolor{oursrow}{HTML}{E9F1FB}
\usepackage{algorithm}
\usepackage{algpseudocode}
\usepackage{amsthm}
\newtheorem{proposition}{Proposition}
\AtEndPreamble{%
  \crefname{algorithm}{Alg.}{Algs.}\Crefname{algorithm}{Algorithm}{Algorithms}%
  \crefname{proposition}{Prop.}{Props.}\Crefname{proposition}{Proposition}{Propositions}%
}

\usepackage{placeins}
\definecolor{cvprblue}{rgb}{0.21,0.49,0.74}
\usepackage[pagebackref,breaklinks,colorlinks,allcolors=cvprblue]{hyperref}

\def\confName{CVPR}
\def\confYear{2027}

\title{Eco3R: Pinned-Key Inference for Efficient and Streamable Feed-Forward 3D Reconstruction from Long Videos}

\author{Hongsen Cao \qquad Shanxin Yuan \qquad Mona Jaber \qquad Ahmed M. Abdelmoniem\\
School of Electronic Engineering and Computer Science\\
Queen Mary University of London, London, UK\\
{\tt\small hong.cao, shanxin.yuan, m.jaber, ahmed.sayed@qmul.ac.uk}
}

\begin{document}
\twocolumn[{%
\renewcommand\twocolumn[1][]{#1}%
\maketitle
\begin{center}
  \centering
  \captionsetup{type=figure,skip=4pt}
  \includegraphics[width=\textwidth]{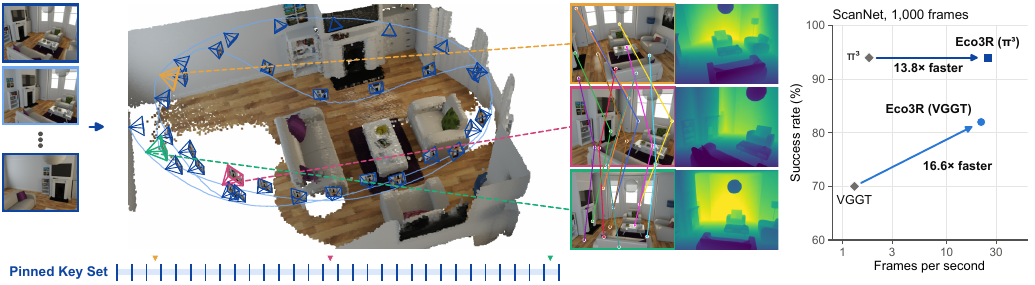}
  \captionof{figure}{\textbf{Eco3R reconstructs every frame of a long video with 31 uniformly sampled frames as the only keys.}
  Eco3R on the $\pi^3$ backbone processes all 1,676 frames of the whiteroom video in NRGBD~\cite{nrgbd} (left) and predicts the camera pose and dense geometry of every frame in one coordinate frame (middle). Dark blue marks the 31 pinned frames of the Pinned Key Set, their camera frustums, and their positions on the timeline (bottom), while light blue marks other frames and the camera trajectory. The panels beside the reconstruction show the input images and predicted depth maps of three frames that are not pinned. Triangles on the timeline mark these frames. In the input images, lines of the same color connect pixels corresponding to the same 3D point under the predicted depth and camera poses. The plot on the right compares Eco3R with the full VGGT and $\pi^3$.}
  \label{fig:scene}
\end{center}
}]
\begin{abstract}
Feed-forward multi-view Transformers such as VGGT recover camera poses and scene geometry in a single forward pass, but their cross-frame attention scales quadratically with the number of frames, while reconstruction accuracy degrades on long videos.
Most existing approaches either approximate attention across all frames under a fixed computational budget or select a small subset of key frames based on their content.
We hypothesize that key frames in long videos can be selected independently of video content. Experiments with VGGT and $\pi^3$ support this hypothesis, showing that a small set of uniformly sampled frames can serve as shared keys for all frames. We present Eco3R, a training-free inference framework that constructs a Pinned Key Set of 31 uniformly sampled frames. A pin pass stores their tokens in a Scene Cache, which the remaining frames access in independent batches. This reproduces joint inference with the pinned keys exactly, while maintaining constant computational cost per frame.
An Online Stream updates the pinned frames during recording. Six intermediate global blocks of VGGT align the frames and form a reusable alignment module. The Alignment Score uses their attention patterns to detect alignment  failures, while Alignment Refocusing mitigates these failures by fine-tuning only these blocks (3.6 GPU hours).
Eco3R is compatible with multiple backbones, including VGGT and $\pi^3$, achieving 6.9--16.6$\times$ speedups while maintaining or improving success rates (70\% to 82\% for VGGT). It reconstructs up to 4,000 frames on a single A100 GPU and streams at 18.6 fps with the lowest online absolute trajectory error.
\end{abstract}%
\begin{figure*}[t]
  \centering
  \includegraphics[width=0.96\textwidth]{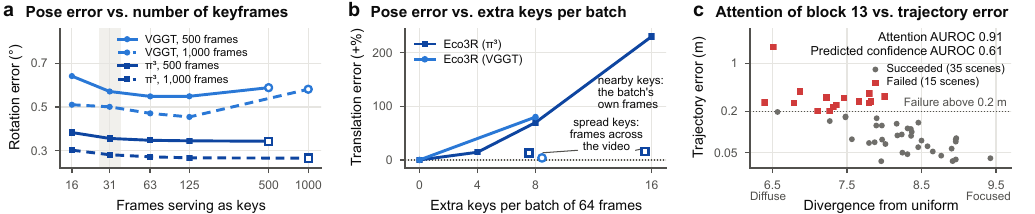}
  \caption{\textbf{Three insights into cross-frame attention on long videos.}
  (a) Median relative rotation error across 50 ScanNet scenes vs. the number of uniformly sampled keyframes, with all frames used as queries.
  (b) Increase of the median relative translation error at 500 frames when batches of 64 consecutive frames add 4, 8, or 16 of their own frames to 31 uniformly sampled keys.
  (c) Trajectory error of the full VGGT at 500 frames against the 90th percentile of the divergence of block 13 from uniform attention, one point per scene.}
  \label{fig:insights}
\end{figure*}
\section{Introduction}
\label{sec:intro}

Feed-forward networks now recover 3D scenes directly from images~\cite{dust3r,mast3r,vggt}. Given a video, the Visual Geometry Grounded Transformer (VGGT)~\cite{vggt} predicts the camera pose and dense geometry of every frame in one forward pass, and later models such as $\pi^3$~\cite{pi3}, VGGT-$\Omega$~\cite{vggtomega}, and Depth Anything 3 (DA3)~\cite{da3} work the same way. Their global blocks make this possible, since cross-frame attention lets each view draw on all other views.

We study how such a model can reconstruct every frame of a long video in one coordinate frame at a constant cost per frame. Long videos challenge these models in two ways. First, the cost of cross-frame attention grows quadratically with the number of frames, and its memory grows proportionally with every frame. A 2,000-frame video at 518 $\times$ 392 resolution requires $>$80\,GB GPU memory, so long videos are split into chunks or subsampled, while subsampling leaves most frames unreconstructed. Second, accuracy suffers even when the video can fit in memory. Our evaluations show that, at 500 frames, the full VGGT keeps its trajectory error within a 0.2\,m tolerance in only 70\% of the 50 ScanNet scenes~\cite{scannet}. Its predicted confidence flags failures only marginally better than chance (\cref{fig:insights}c), leaving users unable to reliably distinguish trustworthy results from failures.

Most existing remedies approximate cross-frame attention over all frames under a fixed computational budget, but each sacrifices some aspect of this goal. Streaming variants~\cite{streamvggt,retrievevggt,infinitevggt} process each frame once as it arrives and freeze its tokens, but early errors persist. They maintain reconstruction errors within the specified tolerance for at most half the scenes (\cref{tab:main}), while InfiniteVGGT incorrectly positions walls in two long videos (\cref{fig:qual}). Chunked pipelines~\cite{vggtlong,laser} process one window at a time, leaving drift to accumulate across successive windows unless corrected by loop closure. Token merging~\cite{fastvggt} processes all frames in one pass but runs out of memory at 2,000 frames in most scenes. Methods that give each frame a few keys select them either at fixed intervals~\cite{kvtracker,sailrecon,spark3r} or based on content, such as viewpoint diversity~\cite{gotohunt,divpart} or frame retrieval~\cite{retrievevggt}. Two questions remain open: whether a long video needs attention across all frames, and whether the keys selected for each frame must be content-dependent.

We hypothesize that the keys of a long video can be selected independently of video content. A shared context of a few dozen frames, uniformly sampled across the video, can provide keys for every frame, eliminating the need for direct attention between the remaining frames. Our experiments on long indoor videos with VGGT and $\pi^3$ support this hypothesis. First, with 31 to 125 uniformly sampled frames as the only keys, VGGT estimates relative rotations more accurately than with all frames, consistent with surplus keys diluting the attention~\cite{gotohunt,divpart} (\cref{fig:insights}a). With 31 keys, $\pi^3$ maintains the trajectory error (\cref{tab:main}). Second, keys chosen by content perform worse than uniformly sampled keys. Keys from farthest-point sampling on place-recognition descriptors, as done in Good Token Hunting~\cite{gotohunt}, raise the trajectory error of $\pi^3$ by 9.3\% (\cref{tab:ablation}), and the 31 most-attended frames raise the relative rotation error of VGGT over sixfold (Sec.~S6). Third, keys outside the shared set provide no additional benefit. Extra keys from nearby frames degrade the relative poses (\cref{fig:insights}b), and prior work finds them poor substitutes for spread keys~\cite{gotohunt,framevggt}.

If a single shared context suffices, videos of arbitrary length can be reconstructed with constant computational cost per frame, while preserving exact attention over the shared context.
We propose Eco3R, an inference framework with three components that make it Exact, Cached, and Online. A Pinned Key Set of 31 uniformly sampled frames provides the sole keys, allowing every frame to access the entire scene in a common coordinate frame (\cref{fig:scene}). The remaining frames can then be processed in independent batches, exactly reproducing joint inference under this attention scheme. Following keyframe tracking~\cite{kvtracker}, a pin pass computes and stores the pinned-frame context in a Scene Cache once, which is then reused by every batch. During recording, an Online Stream keeps the pinned frames spread over the whole observed history and caps their number, so that the cost per frame stays constant as the video grows.

Beyond efficiency, we address the failures of VGGT on long videos. Prior layer analyses locate frame alignment in a band of intermediate global blocks~\cite{avggt,regimevggt,saf3r}. We find that in failed scenes, the attention of block 13 in this band stays closer to uniform, which separates failures from successes with an AUROC of 0.91 (\cref{fig:insights}c). This finding motivates our Alignment Score, which reads this deviation to flag failures, and our Alignment Refocusing, which fine-tunes only the band to reduce them.

When running offline, Eco3R reconstructs as many scenes as its backbones, while at a fraction of their cost (\cref{fig:tradeoff}). With Alignment Refocusing, Eco3R (VGGT) raises VGGT's success rate on ScanNet from 70\% to 82\% while running 8.5$\times$ faster on 500 frames and can handle 4,000 frames on one A100 GPU. Without training, Eco3R ($\pi^3$) matches $\pi^3$'s success rate on 1,000 frames while running 13.8$\times$ faster
(\cref{tab:main,tab:scale}). On four other indoor datasets, it has the lowest absolute trajectory error of all chunked and streaming methods (\cref{tab:heldout}).

Our contributions are threefold: 1)~an observation that a small set of content-agnostic, uniformly sampled keys outperforms content-based key selection and yields lower relative rotation errors for VGGT than using all frames; 2)~Eco3R, a training-free inference framework that computes exact attention over these keys at constant per-frame cost, supports online streaming, and generalizes across backbones, including VGGT, $\pi^3$, VGGT-$\Omega$, and DA3; 3)~a reusable alignment module, with Alignment Refocusing improving success rates,
and a training-free Alignment Score detecting alignment failures.

\begin{figure*}[t]
  \centering
  \includegraphics[width=0.96\textwidth]{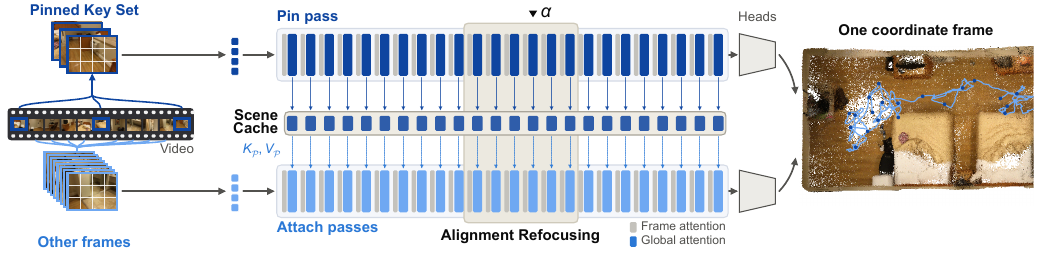}
  \caption{\textbf{Overview of Eco3R.} A pin pass processes the Pinned Key Set of uniformly sampled frames (dark blue) and stores the inputs of every global block in the Scene Cache. In attach passes, the other frames (light blue) read the keys $K_{\mathcal{P}}$ and values $V_{\mathcal{P}}$ only from this Scene Cache. Alignment Refocusing fine-tunes blocks 10--15, while the Alignment Score $\alpha$ is computed from block 13.}
  \label{fig:method}
\end{figure*}
\section{Related work}
\label{sec:related}

\paragraph{Long-video inference}
Acceleration methods for multi-view transformers~\cite{fast3r,vggt,pi3,mapanything,da3} merge tokens~\cite{fastvggt,litevggt,httm} as in vision transformers~\cite{tome}, subsample keys and values on a spatial grid~\cite{avggt,regimevggt}, or skip weak interactions~\cite{sparsevggt,saf3r}, all in one pass over every frame. VGG-T$^3$~\cite{vggttt} retrains the attention to read a fixed-size memory, while FlashVGGT~\cite{flashvggt} retrains it to read compact descriptors of all frames and a few keyframes. Streaming methods keep a spatial memory~\cite{spann3r,long3r}, a learned state~\cite{cut3r,ttt3r}, a multi-layer memory~\cite{must3r}, or a cache of past keys and values~\cite{streamvggt,stream3r,longstream}, which later variants bound by eviction, frame selection, or retrieval~\cite{infinitevggt,xstreamvggt,framevggt,retrievevggt,ovggt,regvggt,evict3r,streamcachevggt,incvggt}. Chunked pipelines join overlapping windows, offline~\cite{vggtlong} or incrementally~\cite{vggtslam,vggtslam2}, and LASER~\cite{laser} aligns them by rescaling each depth layer.

\paragraph{Shared keys}
Methods in which many frames read a few keys come closest to our hypothesis. KV-Tracker~\cite{kvtracker} caches the keys and values of keyframes for real-time tracking. SAIL-Recon~\cite{sailrecon} fine-tunes VGGT so that frames read the cached keys and values of 50 to 100 uniformly sampled anchor frames, while Spark3R~\cite{spark3r} keeps those of about 40 frames at a temporal stride and finds them slightly better than random tokens. In KV-Tracker and SAIL-Recon, each frame also reads its own tokens. Good Token Hunting~\cite{gotohunt} selects one key set for all queries by farthest-point sampling on place-recognition descriptors, while DA-VGGT~\cite{divpart} partitions the frames into chunks of diverse views that share one anchor frame.

\paragraph{Long-context attention}
In transformers for text, scaling the logits by the logarithm of the input length restores the focus that the softmax loses over long inputs~\cite{chiang2022,ssmax}, while StreamingLLM~\cite{streamingllm} keeps a few initial tokens as permanent keys next to recent ones. Multi-view models lose focus in the same way: the cross-frame attention of VGGT collapses toward rank one over hundreds of frames~\cite{fastvggt,collapse}, and redundant views dilute it~\cite{divpart}. Layer analyses find that middle global blocks align the views~\cite{avggt,regimevggt,saf3r} and attend selectively~\cite{regvggt,evict3r}. Partial fine-tuning has adapted the image encoder with low-rank adapters for finer geometry~\cite{fin3r,lora} and the middle blocks of VGGT for dynamic scenes~\cite{page4d}. Uncertainty estimates use the often overconfident predicted confidence~\cite{calibaudit}, trained adapters~\cite{geocond}, or perturbed reruns~\cite{revar3r}, while VGGT-SLAM 2.0~\cite{vggtslam2} verifies loop closures with one attention layer.
\section{Method}
\label{sec:method}
\label{sec:analysis}

Eco3R organizes VGGT-style inference around one shared context of pinned frames (\cref{fig:method}). It computes the shared context once, reads it exactly, and updates it online. For VGGT, an alignment module refocuses the blocks that align the frames and reads their attention to flag failures (\cref{sec:method_refocus}). Analyses in this section use the 50 ScanNet scenes at 500 frames unless otherwise noted (\cref{sec:exp_setup}).

Consider a video of $N$ frames, each represented by $T$ tokens of dimension $C$. The network alternates between $L$ frame blocks that attend to tokens within each frame (frame attention) and $L$ global blocks that attend to tokens across frames (cross-frame attention).

\subsection{Cross-frame attention}
\label{sec:method_attn}

For a single attention head in a global block, let $X_i \in \mathbb{R}^{T \times C}$ denote the tokens of frame $i$ and $\mathbf{X}$ the concatenated tokens of all frames. The attention output for frame $i$ is
\begin{equation}
  O_i = \mathrm{softmax}\Big(\frac{Q_i K^{\top}}{\sqrt{d}}\Big) V,
  \label{eq:global}
\end{equation}
where $Q_i = X_i W_Q$, $K = \mathbf{X} W_K$, and $V = \mathbf{X} W_V$, with $W_Q$, $W_K$, and $W_V$ denoting the query, key, and value projection matrices, respectively, and $d$ the dimension of each attention head.

Let $H_i^{(l)}$ be the tokens of frame $i$ entering global block ${l \in \{0, \dots, L-1\}}$. For ${l < L-1}$, a joint pass computes
\begin{equation}
  H_i^{(l+1)} = F_{l+1}\Big( G_l\big( H_i^{(l)} ;\, H^{(l)} \big) \Big),
  \label{eq:recursion}
\end{equation}
where $H^{(l)}$ stacks the tokens of all frames, $F_{l+1}$ is the next frame block, and $G_l(H_i^{(l)} ;\, H^{(l)})$ is global block $l$ with queries from $H_i^{(l)}$ and keys and values from $H^{(l)}$. The initial tokens $H_i^{(0)}$ depend only on frame $i$, and all VGGT prediction heads (except the camera head) process each frame independently.

For a single query and attention head, let $\mathbf{a} = (a_1, \dots, a_n)$ denote the attention weights over $n$ keys, and let $\mathbf{u}$ denote the uniform distribution. We measure the deviation of $\mathbf{a}$ from uniformity as
\begin{equation}
  \mathrm{KL}(\mathbf{u} \,\|\, \mathbf{a}) = -\log n - \frac{1}{n} \sum_{j=1}^{n} \log a_j ,
  \label{eq:div}
\end{equation}
a Kullback--Leibler divergence that increases as more keys receive negligible attention weights. The attention share of one frame on another is the attention that its queries give to all keys of the other, averaged over queries and heads.

\begin{figure}[t]
  \centering
  \includegraphics[width=\linewidth]{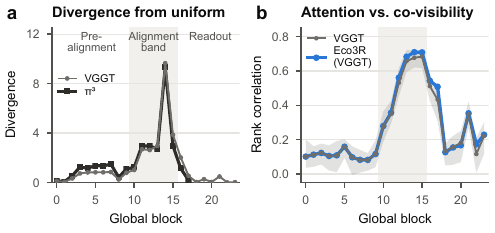}
  \caption{\textbf{The alignment band.}
  (a) Divergence from uniform attention per global block.
  (b) Rank correlation between the attention share of two distant frames and their co-visibility,
  with the interquartile range of VGGT in gray.}
  \label{fig:band}
\end{figure}

\subsection{The alignment band}
\label{sec:method_band}

 The alignment band is a subset of global blocks characterized by high divergence and high correlation.
To measure attention divergence, we evaluate \cref{eq:div} for each global block using three 100-frame sequences from 7-Scenes~\cite{sevenscenes}.
Both VGGT and $\pi^3$ exhibit a similar trend: the median divergence remains near or below 1 across blocks 0--9, peaks around block 14, and falls below 1 by block 17 (\cref{fig:band}a).
To measure correlation, we uniformly sample 31 frames from each of the 50 ScanNet scenes and measure the correlation between attention share and surface co-visibility for pairs of distant frames separated by at least four pinned frames. Co-visibility is defined as the overlap between the surfaces observed by the two frames.
For VGGT, the rank correlation remains at or below 0.16 across blocks 0--9, rising to 0.66--0.68 in blocks 13--15 (\cref{fig:band}b). This suggests that these blocks relate frames by their content.

Based on the divergence and correlation analyses, we designate blocks 10--15 as the alignment band and the subsequent blocks as the readout blocks.

\subsection{Pinned Key Set}
\label{sec:method_pks}

Using only 31 to 125 uniformly sampled frames as keys reduces the median relative rotation error of VGGT (\cref{fig:insights}a). At 1,000 frames, using 31 keys reduces the error from 0.581\textdegree{} to 0.501\textdegree{}. At 500 frames, using all 500 frames as keys increases the error by 7.2\% relative to using 125 keys (0.549\textdegree{}). These results suggest that excessive keys may dilute attention, particularly when their number exceeds the 24-frame training limit~\cite{vggt} by a factor of 10--20.

Let $\mathcal{P} = \{p_1, \dots, p_M\}$ denote the indices of $M$ pinned frames, sampled uniformly in time as ${p_k = 1 + \mathrm{round}\big((k-1)(N-1)/(M-1)\big)}$. We restrict the keys and values of every global block to the pinned frames:
\begin{equation}
  O_i = \mathrm{softmax}\Big(\frac{Q_i K_{\mathcal{P}}^{\top}}{\sqrt{d}}\Big) V_{\mathcal{P}},
  \label{eq:pks}
\end{equation}
where $K_{\mathcal{P}}$ and $V_{\mathcal{P}}$ contain the rows of $K$ and $V$ that belong to the pinned frames. We set ${M = 31}$, the smallest tested key set size that preserves the success rates of both VGGT and  $\pi^3$ on the evaluation scenes.

Let $H_{\mathcal{P}}^{(l)}$ denote the stacked tokens of the pinned frames entering global block $l$. With pinned keys, the joint pass in \cref{eq:recursion} becomes
\begin{equation}
  H_i^{(l+1)} = F_{l+1}\Big( G_l\big( H_i^{(l)} ;\, H_{\mathcal{P}}^{(l)} \big) \Big).
  \label{eq:pass}
\end{equation}
Since the pinned frames attend only to one another, a pin pass can process them alone and store $H_{\mathcal{P}}^{(l)}$ at every global block. An attach pass then processes the remaining frames in batches and computes their tokens with \cref{eq:pass}, forming the keys and values of every global block from the stored $H_{\mathcal{P}}^{(l)}$. Since each of these frames attends only to the pinned keys, the frames within a batch do not attend to one another.

\begin{proposition}[Exact decomposition]\leavevmode
\label{prop:exact}
Consider a joint pass over all $N$ frames using \cref{eq:pks} in every global block, with the first frame pinned. Under exact arithmetic, a pin pass over $\mathcal{P}$ followed by attach passes over the remaining frames, processed in batches of arbitrary size and order, yields identical tokens and predictions to the joint pass for every frame, provided that the camera head processes the same frames in both cases (proof in Sec.~S6).
\end{proposition}

As a result, all predictions share one coordinate frame. In each global block, the number of query-key products per head becomes
\begin{equation}
  \underbrace{M^2T^2\vphantom{(N - M)}}_{\text{pin pass}} + \underbrace{(N - M)\,M\,T^2}_{\text{attach passes}} \;=\; N M T^2,
  \label{eq:cost}
\end{equation}
which scales linearly with $N$ and is a factor of $N/M$ smaller than the $N^2T^2$ of full attention (16$\times$ at 500 frames and 129$\times$ at 4,000 frames).

Attached frames attend only to the pinned keys, as including additional nearby keys degrades camera pose accuracy (\cref{fig:insights}b). Adding 4, 8, or 16 frames from each batch to the key set increases the relative translation error of Eco3R ($\pi^3$) by 14\%, 70\%, and 230\%, respectively. With 16 additional frames, the success rate drops from 94\% to 78\%. Adding 8 nearby keys increases the relative translation error of Eco3R (VGGT) by 80\%, whereas adding eight keys sampled from across the whole video increases it by 4\% for VGGT and 13\% for $\pi^3$. Our analysis suggests that this degradation is primarily attributable to near-duplicate views  (Sec.~S7).

\begin{figure*}[t]
  \centering
  \includegraphics[width=0.96\textwidth]{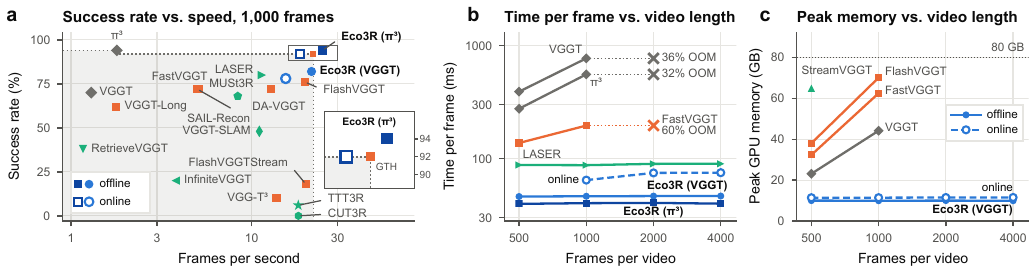}
  \caption{\textbf{Accuracy and cost on ScanNet.}
  (a) Success rate vs. speed (fps) at 1,000 frames. The dotted line denotes the Pareto front of competing methods, with the region below it shaded and the corner shown in an enlarged inset.
  GTH is Good Token Hunting. (b) Median time per frame and (c) peak inference memory against video length.}
  \label{fig:tradeoff}
\end{figure*}

\subsection{Scene Cache and Online Stream}
\label{sec:method_online}
\label{sec:method_cache}

The Scene Cache ${\mathcal{C} = \{ H_{\mathcal{P}}^{(l)} \}_{l=0}^{L-1}}$ stores the $L\,M\,T$ tokens that the pin pass computes at the global blocks.  For VGGT, the cache occupies 1.5\,GB in bfloat16 regardless of the video length. Each attach batch moves its outputs off the GPU, so the memory of a pass depends only on $M$ and the batch size.

The Online Stream updates the pinned frames during recording (Algorithm~S1). Every $\kappa$ frames, the newest frame becomes a candidate key. Let ${c_m = 1 + m\kappa}$ denote the $m$-th candidate. Pass $m$ pins at most $M_{\max}$ of the candidates $c_0, \dots, c_m$, sampled uniformly:
\begin{equation}
  \mathcal{P}_m = \big\{ c_{r_j} \big\}_{j=0}^{M_{\max}-1}, \qquad r_j = \mathrm{round}\Big(\frac{j\,m}{M_{\max}-1}\Big).
  \label{eq:thin}
\end{equation}
With a lag of $\tau$ passes, pass $m$ attaches and emits the remaining frames up to $c_{m-\tau}$. Each emitted frame thus attends to keys from the whole observed history and to at most $\tau + 1$ later candidates.

Every key set contains the first frame, where VGGT fixes the coordinate frame, but each pass normalizes the scale according to its own frames. To keep one scale across passes, each pass rescales its depths and camera translations by the median ratio of the emitted depth to its own predicted depth over its pinned frames  that earlier passes have emitted (Sec.~S1). For $\pi^3$, which has no reference frame, a rigid transform further aligns each pass with the poses emitted earlier. The camera head of VGGT processes only the frames of each pass. Each pass processes at most $M_{\max}$ keys, which bounds its memory, and every frame is emitted within ${(\tau + 1)\,\kappa}$ frames of its arrival. We use ${M_{\max} = 32}$ and ${\kappa = 32}$ (16 at 500 frames), with ${\tau = 2}$ for VGGT and ${\tau = 0}$ for $\pi^3$.

\subsection{Alignment Refocusing and Alignment Score}
\label{sec:method_refocus}
\label{sec:method_score}

We trace the remaining failures of VGGT to the alignment band. With 31 uniformly sampled keys, the frozen VGGT succeeds in 72\% of the scenes. Its failures occur in large rooms. The ground-truth extent of the camera path separates them from the successes with an AUROC of 0.86. Failed scenes contain at least as many distant co-visible pairs among their pinned frames as successful scenes, but blocks 12--15 of the band assign these pairs a sixth less attention. Scaling the attention logits by 1.15 reduces the rotation error of the full VGGT by 2\% and 18\% at 500 and 1,000 frames, respectively (Sec.~S7), but it sharpens all attention uniformly.

These findings motivate our Alignment Refocusing, which fine-tunes only the band (76M parameters, or 6.5\% of VGGT) to align distant co-visible frames. The training samples follow the structure of inference, so that the band learns to align frames through pinned keys. From a whole scan or a random contiguous half of it, we pin 24 uniformly sampled frames and draw 24 other random frames that attend only to these keys. We minimize the camera and depth losses of VGGT~\cite{vggt} with respect to the weights of the band (Sec.~S6). Training on 195 ScanNet scans of rooms outside our evaluation set takes 2,000 AdamW~\cite{adamw} steps with a learning rate of $10^{-5}$ (3.6 GPU hours).

The band also indicates failures. For the full VGGT at 500 frames, the 90th percentile of the divergence in \cref{eq:div}, computed over 2,048 random queries and all heads of block 13, separates failed scenes from successful ones with an AUROC of 0.91 (\cref{fig:insights}c). We define the Alignment Score $\alpha$ as this 90th percentile computed in the pin pass, where the divergence costs one dot product per query (Sec.~S8). The score is available before any attach pass and requires neither training nor ground truth. A low score indicates a likely failure (\cref{sec:exp_score}).

\section{Experiments}
\label{sec:experiments}

\begin{table*}[t]
  \caption{\textbf{Long ScanNet videos on one A100 GPU.}
  Units are meters for absolute trajectory error (ATE), degrees for relative rotation error (RPE), centimeters for point accuracy (Acc.), seconds per scene for runtime, and GB for peak inference memory (maximum across scenes).
  Shaded rows are Eco3R (ours), while bold values denote  the best results in each group.
  ``Failed'' indicates a success rate of at most 20\%, with the rate shown in parentheses.
  $^{\dagger}$Trains or fine-tunes network weights. $^{\ddagger}$Full pipeline (inference 74\,s and 151\,s).}
  \label{tab:main}
  \centering
  \scriptsize
  \setlength{\tabcolsep}{3.9pt}
  \renewcommand{\arraystretch}{0.88}
  \begin{tabular}{@{}l rrrrrr c rrrrrr@{}}
  \toprule
  & \multicolumn{6}{c}{500 frames} & & \multicolumn{6}{c}{1,000 frames} \\
  \cmidrule(lr){2-7} \cmidrule(lr){9-14}
  Method & Success$\uparrow$ & ATE$\downarrow$ & RPE$\downarrow$ & Acc.$\downarrow$ & Time$\downarrow$ & Mem.$\downarrow$ & & Success$\uparrow$ & ATE$\downarrow$ & RPE$\downarrow$ & Acc.$\downarrow$ & Time$\downarrow$ & Mem.$\downarrow$ \\
  \midrule
VGGT (full model)~\cite{vggt} & 70 & 0.111 & 0.589 & 1.97 & 196 & 23.1 &  & 70 & 0.130 & 0.581 & 2.26 & 772 & 44.2 \\
$\pi^3$ (full model)~\cite{pi3} & 94 & 0.054 & 0.343 & 0.94 & 138 & 24.4 &  & 94 & 0.058 & 0.265 & 0.90 & 557 & 43.4 \\
\midrule
\multicolumn{14}{@{}l}{\textit{Offline}} \\
FastVGGT~\cite{fastvggt} & 74 & 0.098 & 0.598 & 1.95 & 68.8 & 32.3 &  & 72 & 0.100 & 0.521 & 2.00 & 196 & 62.5 \\
FlashVGGT~\cite{flashvggt}$^{\dagger}$ & 76 & 0.121 & 0.563 & 1.88 & \textbf{19.7} & 38.0 &  & 76 & 0.118 & 0.426 & 1.80 & 50.6 & 70.1 \\
FlashVGGTStream~\cite{flashvggt}$^{\dagger}$ & 26 & 0.307 & 2.013 & 8.70 & 20.5 & 18.0 &  & failed (18) & — & — & — & — & — \\
VGG-T$^3$~\cite{vggttt}$^{\dagger}$ & failed (14) & — & — & — & — & — &  & failed (10) & — & — & — & — & — \\
Good Token Hunting~\cite{gotohunt} ($\pi^3$) & 90 & 0.062 & 0.358 & 1.06 & 23.0 & 25.2 &  & 92 & 0.067 & \textbf{0.280} & 1.03 & 45.4 & 44.2 \\
SAIL-Recon~\cite{sailrecon}$^{\dagger}$ & 70 & 0.118 & 0.492 & 1.95 & 104 & 13.1 &  & 72 & 0.108 & 0.405 & 2.01 & 199 & 13.1 \\
DA-VGGT~\cite{divpart} & 72 & 0.118 & 0.671 & 2.03 & 38.7 & 13.1 &  & 72 & 0.119 & 0.580 & 2.08 & 78.0 & 13.5 \\
VGGT-Long~\cite{vggtlong}$^{\ddagger}$ & 68 & 0.129 & 0.376 & 2.18 & 263 & 20.4 &  & 62 & 0.164 & 0.328 & 2.70 & 547 & 20.5 \\
\rowcolor{oursrow} Eco3R (VGGT)$^{\dagger}$ & 82 & 0.090 & 0.528 & 1.48 & 23.1 & 10.0 &  & 82 & 0.093 & 0.434 & 1.53 & 46.5 & 10.0 \\
\rowcolor{oursrow} Eco3R ($\pi^3$) & \textbf{94} & \textbf{0.052} & \textbf{0.356} & \textbf{1.00} & 19.9 & \textbf{9.9} &  & \textbf{94} & \textbf{0.060} & 0.281 & \textbf{1.00} & \textbf{40.4} & \textbf{9.9} \\
\midrule
\multicolumn{14}{@{}l}{\textit{Online}} \\
CUT3R~\cite{cut3r}$^{\dagger}$ & failed (2) & — & — & — & — & — &  & failed (0) & — & — & — & — & — \\
TTT3R~\cite{ttt3r} & failed (20) & — & — & — & — & — &  & failed (6) & — & — & — & — & — \\
MUSt3R~\cite{must3r}$^{\dagger}$ & 64 & 0.148 & 0.620 & 3.07 & 69.3 & 9.6 &  & 68 & 0.149 & 0.499 & 3.06 & 116 & 11.3 \\
StreamVGGT~\cite{streamvggt}$^{\dagger}$ & 48 & 0.210 & 1.191 & 6.20 & 456 & 64.9 &  & \multicolumn{6}{c}{out of memory} \\
RetrieveVGGT~\cite{retrievevggt} & 38 & 0.235 & 2.534 & 7.26 & 407 & 12.6 &  & 38 & 0.281 & 2.552 & 7.64 & 856 & 12.7 \\
InfiniteVGGT~\cite{infinitevggt} & 30 & 0.262 & 2.627 & 6.83 & 127 & 14.9 &  & failed (20) & — & — & — & — & — \\
LASER~\cite{laser} & 84 & 0.107 & \textbf{0.316} & 2.05 & 43.9 & \textbf{6.6} &  & 80 & 0.120 & \textbf{0.232} & 2.61 & 87.0 & \textbf{6.6} \\
VGGT-SLAM~\cite{vggtslam} & 50 & 0.192 & 0.468 & 3.81 & 41.9 & 10.8 &  & 48 & 0.183 & 0.448 & 5.26 & 90.4 & 12.1 \\
\rowcolor{oursrow} Eco3R (VGGT)$^{\dagger}$ & 78 & 0.102 & 0.540 & 1.67 & 45.8 & 11.4 &  & 78 & 0.099 & 0.449 & 1.69 & 64.6 & 11.5 \\
\rowcolor{oursrow} Eco3R ($\pi^3$) & \textbf{90} & \textbf{0.061} & 0.368 & \textbf{1.22} & \textbf{37.0} & 11.0 &  & \textbf{92} & \textbf{0.061} & 0.280 & \textbf{1.13} & \textbf{53.8} & 11.0 \\
  \bottomrule
\end{tabular}
\end{table*}

\label{sec:exp_setup}
\paragraph{Benchmarks} We follow the long-video protocol of FastVGGT~\cite{fastvggt} on its 50 ScanNet scenes~\cite{scannet}, using its frame budgets of 500 and 1,000 frames, to which we add budgets of 2,000 and 4,000 frames. Videos shorter than the budget keep all their frames (Sec.~S3). We report the absolute trajectory error (ATE), the relative rotation error (RPE) and translation error between consecutive frames, the accuracy (Acc.) and completeness (Comp.) of point clouds, and the absolute relative depth error (AbsRel).
A reconstruction is successful when its ATE is below 0.2\,m.
We report scene-level medians, with all relative changes calculated from these median values. We assess significance with Wilcoxon signed-rank tests over scenes. We further evaluate point clouds and trajectories on 7-Scenes~\cite{sevenscenes}, NRGBD~\cite{nrgbd}, TUM RGB-D~\cite{tum}, and Replica~\cite{replica}, as well as trajectories on KITTI odometry~\cite{kitti}.

\paragraph{Baselines and implementation}
Besides the full VGGT, $\pi^3$, VGGT-$\Omega$, and DA3, we compare with 21 state-of-the-art methods.
Unless noted otherwise, all methods run with their official code and weights on 80\,GB A100 GPUs (Sec.~S2).

\subsection{Comparison with State-of-the-art}
\label{sec:exp_main}

\paragraph{Offline}
\label{sec:exp_offline}
Eco3R preserves or improves the success rate of each backbone while substantially reducing computational cost  (\cref{tab:main}, \cref{fig:tradeoff}a)\footnote{Additional results are provided in the supplementary material.}.
At 500 frames, Eco3R (VGGT) improves VGGT's success rate from 70\% to 82\% (95\% confidence interval for the gain: 4 to 22 points) while reducing all errors, including trajectory error by 19\%, relative translation error by 26\%, and accuracy error by 25\% (Sec.~S3). It achieves speedups of  8.5$\times$ and 16.6$\times$  over VGGT at 500 and 1,000 frames, respectively. Eco3R ($\pi^3$) preserves the success rate and trajectory accuracy of the full $\pi^3$ model while achieving speedups of 6.9$\times$ and 13.8$\times$ at 500 and 1,000 frames, respectively. At 1,000 frames, Eco3R consumes only 7\% and 6\% of the energy required by the full $\pi^3$ and VGGT, respectively (\cref{tab:eff}). Eco3R (DA3) preserves the trajectory accuracy of the full DA3 model at both 500 and 1,000 frames, achieving a 4.9$\times$ speedup at 500 frames.

\begin{table}[t]
  \caption{\textbf{Thousands of frames on ScanNet.} Units and notation follow \cref{tab:main}, with time per frame in ms. Scenes that exceed 80\,GB of memory count as failures, and ATE and time are medians over the remaining scenes.}
  \label{tab:scale}
  \centering
  \footnotesize
  \setlength{\tabcolsep}{2.6pt}
  \renewcommand{\arraystretch}{0.91}
  \resizebox{\linewidth}{!}{\begin{tabular}{@{}l rrrr c rrrr@{}}
  \toprule
  & \multicolumn{4}{c}{2,000 frames} & & \multicolumn{4}{c}{4,000 frames} \\
  \cmidrule(lr){2-5} \cmidrule(lr){7-10}
  Method & Succ.$\uparrow$ & ATE$\downarrow$ & ms/fr.$\downarrow$ & Mem.$\downarrow$ & & Succ.$\uparrow$ & ATE$\downarrow$ & ms/fr.$\downarrow$ & Mem.$\downarrow$ \\
  \midrule
VGGT~\cite{vggt} & 42 & 0.122 & 846 & $>$80 &  & 42 & 0.122 & 846 & $>$80 \\
$\pi^3$~\cite{pi3} & 60 & 0.050 & 641 & $>$80 &  & 60 & 0.050 & 641 & $>$80 \\
\midrule
\multicolumn{10}{@{}l}{\textit{Offline}} \\
FlashVGGTStream~\cite{flashvggt}$^{\dagger}$ & failed (14) & — & — & — &  & failed (14) & — & — & — \\
VGGT-Long~\cite{vggtlong}$^{\ddagger}$ & 56 & 0.169 & 566 & 20.4 &  & 56 & 0.171 & 565 & 20.4 \\
VGG-T$^3$~\cite{vggttt}$^{\dagger}$ & failed (6) & — & — & — &  & failed (6) & — & — & — \\
\rowcolor{oursrow} Eco3R (VGGT)$^{\dagger}$ & 82 & 0.101 & 47 & 10.0 &  & 82 & 0.098 & 47 & 10.0 \\
\rowcolor{oursrow} Eco3R ($\pi^3$) & \textbf{92} & \textbf{0.060} & \textbf{41} & \textbf{9.9} &  & \textbf{90} & \textbf{0.060} & \textbf{40} & \textbf{9.9} \\
\midrule
\multicolumn{10}{@{}l}{\textit{Online}} \\
CUT3R~\cite{cut3r}$^{\dagger}$ & failed (0) & — & — & — &  & failed (0) & — & — & — \\
TTT3R~\cite{ttt3r} & failed (4) & — & — & — &  & failed (4) & — & — & — \\
MUSt3R~\cite{must3r}$^{\dagger}$ & 72 & 0.150 & 142 & 14.4 &  & 72 & 0.150 & 142 & 20.7 \\
RetrieveVGGT~\cite{retrievevggt} & 36 & 0.295 & 866 & 13.0 &  & 34 & 0.295 & 871 & 13.6 \\
InfiniteVGGT~\cite{infinitevggt} & failed (12) & — & — & — &  & failed (12) & — & — & — \\
LASER~\cite{laser} & 74 & 0.140 & 90 & \textbf{6.6} &  & 74 & 0.140 & 90 & \textbf{6.6} \\
\rowcolor{oursrow} Eco3R (VGGT)$^{\dagger}$ & 78 & 0.102 & 75 & 11.5 &  & 78 & 0.103 & 75 & 11.6 \\
\rowcolor{oursrow} Eco3R ($\pi^3$) & \textbf{90} & \textbf{0.064} & \textbf{63} & 11.1 &  & \textbf{92} & \textbf{0.064} & \textbf{62} & 11.1 \\
  \bottomrule
\end{tabular}
}
\end{table}

Eco3R (VGGT) reduces FlashVGGT's trajectory error by 25\% and 21\% at 500 and 1,000 frames while using only 26\% and 14\% of its memory, respectively.
Eco3R (VGGT) succeeds in 82\% of the scenes, whereas SAIL-Recon and DA-VGGT succeed in 70\% to 72\% at both frame lengths. Eco3R (VGGT) reduces their trajectory and accuracy errors by 14\% to 27\% and runs 1.7$\times$ to 4.5$\times$ faster.
Eco3R ($\pi^3$) succeeds more often than Good Token Hunting (\cref{tab:main}), Wang et al.~\cite{sparsevggt}, and SAF3R~\cite{saf3r}, achieving significantly lower trajectory errors, and runs 1.1$\times$ to 6.3$\times$ faster.
Compared with VGGT-Long, Eco3R ($\pi^3$) reduces the trajectory and accuracy errors by 54\% to 64\%.

\begin{table}[t]
  \caption{\textbf{Efficiency on ScanNet} on one A100 GPU. TFLOPs count a multiply-add as two floating-point operations. Latency is the median and first-frame delay when 1,000 frames arrive at 10 per second. Success rate and GPU energy are measured at 1,000 frames. OOM marks out of memory, and other notation follows \cref{tab:main}.}
  \label{tab:eff}
  \centering
  \footnotesize
  \setlength{\tabcolsep}{2.6pt}
  \renewcommand{\arraystretch}{0.91}
  \resizebox{\linewidth}{!}{\begin{tabular}{@{}l r rr rr rr r@{}}
  \toprule
  & Succ.$\uparrow$ & \multicolumn{2}{c}{TFLOPs/frame$\downarrow$} & \multicolumn{2}{c}{FPS$\uparrow$} & \multicolumn{2}{c}{Latency (s)$\downarrow$} & Energy$\downarrow$ \\
  \cmidrule(lr){3-4} \cmidrule(lr){5-6} \cmidrule(lr){7-8}
  Method & (\%) & 500 & 1,000 & 500 & 1,000 & Median & First & (J/frame) \\
  \midrule
VGGT~\cite{vggt} & 70 & 55.6 & 108.9 & 2.6 & 1.3 & 822 & 872 & 223.4 \\
$\pi^3$~\cite{pi3} & 94 & 42.2 & 82.1 & 3.6 & 1.8 & 607 & 656 & 167.0 \\
\midrule
\multicolumn{9}{@{}l}{\textit{Offline}} \\
FastVGGT~\cite{fastvggt} & 72 & 13.7 & 25.1 & 7.3 & 5.1 & 246 & 296 & 51.7 \\
FlashVGGT~\cite{flashvggt}$^{\dagger}$ & 76 & \textbf{4.1} & 5.9 & \textbf{25.3} & 19.8 & 101 & 150 & 14.7 \\
VGG-T$^3$~\cite{vggttt}$^{\dagger}$ & failed (10) & — & — & — & — & — & — & — \\
VGGT-Long~\cite{vggtlong}$^{\ddagger}$ & 62 & 18.6 & 18.7 & 1.9 & 1.8 & 597 & 646 & 72.5 \\
\rowcolor{oursrow} Eco3R (VGGT)$^{\dagger}$ & 82 & 5.6 & 5.6 & 21.6 & 21.5 & 96.4 & 146 & 13.7 \\
\rowcolor{oursrow} Eco3R ($\pi^3$) & \textbf{94} & 4.8 & \textbf{4.8} & 25.1 & \textbf{24.8} & \textbf{90.3} & \textbf{140} & \textbf{11.5} \\
\midrule
\multicolumn{9}{@{}l}{\textit{Online}} \\
CUT3R~\cite{cut3r}$^{\dagger}$ & failed (0) & — & — & — & — & — & — & — \\
TTT3R~\cite{ttt3r} & failed (6) & — & — & — & — & — & — & — \\
StreamVGGT~\cite{streamvggt}$^{\dagger}$ & OOM & 29.2 & OOM & 1.1 & OOM & OOM & OOM & OOM \\
RetrieveVGGT~\cite{retrievevggt} & 38 & 7.7 & 7.8 & 1.2 & 1.2 & 346 & 0.26 & 106.4 \\
InfiniteVGGT~\cite{infinitevggt} & failed (20) & — & — & — & — & — & — & — \\
LASER~\cite{laser} & 80 & \textbf{6.8} & 6.9 & 11.4 & 11.4 & 3.9 & 5.30 & 20.1 \\
\rowcolor{oursrow} Eco3R (VGGT)$^{\dagger}$ & 78 & 9.4 & 6.8 & 10.9 & 15.5 & 9.9 & \textbf{0.09} & 18.4 \\
\rowcolor{oursrow} Eco3R ($\pi^3$) & \textbf{92} & 7.8 & \textbf{5.7} & \textbf{13.5} & \textbf{18.6} & \textbf{3.2} & 0.22 & \textbf{15.1} \\
  \bottomrule
\end{tabular}
}
\end{table}

\paragraph{Online}
\label{sec:exp_online}
Across all sequence lengths, the Online Stream's success rate remains within 4 points of the offline counterpart, with no statistically significant difference in trajectory error in per-scene comparisons (\cref{tab:main,tab:scale}).
At 500 and 1,000 frames, with keys spread over the whole observed history, Eco3R on both backbones reduces the trajectory errors of StreamVGGT, RetrieveVGGT, and InfiniteVGGT by 51\% to 83\%, achieving a lower error than each of them in at least 96\% of the scenes (Sec.~S3).
Eco3R reduces the trajectory error of VGGT-SLAM by 46\% to 68\%. With the same $\pi^3$ weights, Eco3R ($\pi^3$) reduces the trajectory and accuracy errors of LASER by 40\% to 57\%. It sustains 18.6 frames per second (\cref{tab:eff}).

\begin{figure*}[t]
  \centering
  \includegraphics[width=0.9\textwidth]{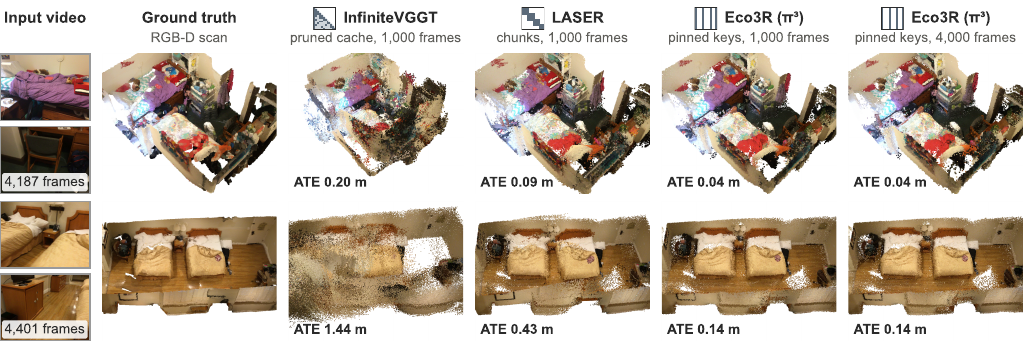}
  \caption{\textbf{Two baselines and Eco3R on two long ScanNet videos}, scene0648\_00 (top) and scene0477\_00 (bottom), each in one oblique view. Icons mark which frames (columns) each frame (rows) attends to: a pruned cache, overlapping chunks, or the same pinned frames.}
  \label{fig:qual}
\end{figure*}

\paragraph{Thousands of frames}
\label{sec:exp_scale}
When running offline on VGGT and $\pi^3$, Eco3R maintains its success rate within 4 points from 500 to 4,000 frames, while keeping its cost per frame and peak memory  constant (\cref{tab:main,tab:scale}, \cref{fig:tradeoff}b, c). Eco3R (VGGT) processes 4,000 frames in three minutes.
Unlike the baselines in \cref{fig:qual}, Eco3R ($\pi^3$) correctly reconstructs all walls  at both 1,000 and 4,000 frames.

\begin{table}[t]
  \caption{\textbf{Four other datasets.} Units and notation follow \cref{tab:main}. $^{\ast}$Stops on 1 of 18 sequences of 7-Scenes and 2 of 12 sequences of TUM RGB-D, and the medians exclude these sequences.}
  \label{tab:heldout}
  \centering
  \footnotesize
  \setlength{\tabcolsep}{2.6pt}
  \resizebox{\linewidth}{!}{\begin{tabular}{@{}l rr c rr c rr c rr@{}}
  \toprule
  & \multicolumn{2}{c}{7-Scenes} & & \multicolumn{2}{c}{NRGBD} & & \multicolumn{2}{c}{TUM RGB-D} & & \multicolumn{2}{c}{Replica} \\
  \cmidrule(lr){2-3} \cmidrule(lr){5-6} \cmidrule(lr){8-9} \cmidrule(lr){11-12}
  Method & Acc. & ATE & & Acc. & ATE & & Acc. & ATE & & Acc. & ATE \\
  \midrule
VGGT~\cite{vggt} & 1.75 & 0.124 & & 1.91 & 0.052 & & 1.43 & 0.076 & & 4.70 & 0.182 \\
$\pi^3$~\cite{pi3} & 1.07 & 0.088 & & 1.27 & 0.030 & & 0.99 & 0.073 & & 1.08 & 0.030 \\
\midrule
\multicolumn{12}{@{}l}{\textit{Offline}} \\
FastVGGT~\cite{fastvggt} & 1.62 & 0.127 & & 1.78 & 0.045 & & 1.51 & 0.079 & & 3.38 & 0.098 \\
VGGT-Long~\cite{vggtlong} & 1.77 & 0.159 & & 1.82 & 0.075 & & 1.93 & 0.127 & & 1.82 & 0.078 \\
\rowcolor{oursrow} Eco3R (VGGT)$^{\dagger}$ & 1.18 & 0.100 & & 1.77 & 0.055 & & 1.10 & 0.091 & & 3.90 & 0.254 \\
\rowcolor{oursrow} Eco3R ($\pi^3$) & \textbf{1.08} & \textbf{0.088} & & \textbf{1.30} & \textbf{0.031} & & \textbf{1.01} & \textbf{0.077} & & \textbf{1.08} & \textbf{0.031} \\
\midrule
\multicolumn{12}{@{}l}{\textit{Online}} \\
CUT3R~\cite{cut3r}$^{\dagger}$ & \multicolumn{2}{c}{failed (0)} & & \multicolumn{2}{c}{failed (0)} & & \multicolumn{2}{c}{failed (0)} & & \multicolumn{2}{c}{failed (0)} \\
TTT3R~\cite{ttt3r} & 4.96 & 0.276 & & \multicolumn{2}{c}{failed (11)} & & \multicolumn{2}{c}{failed (17)} & & \multicolumn{2}{c}{failed (0)} \\
InfiniteVGGT~\cite{infinitevggt} & 3.94 & 0.178 & & 5.47 & 0.172 & & 4.38 & 0.168 & & 9.08 & 0.248 \\
LASER~\cite{laser} & 2.10 & 0.115 & & 2.46 & 0.132 & & 1.87 & 0.128 & & 2.07 & 0.050 \\
VGGT-SLAM~\cite{vggtslam}$^{\ast}$ & 2.31 & 0.134 & & 2.35 & 0.114 & & 2.29 & 0.160 & & 5.13 & 0.104 \\
\rowcolor{oursrow} Eco3R (VGGT)$^{\dagger}$ & \textbf{1.20} & 0.098 & & 2.08 & 0.062 & & 1.34 & \textbf{0.094} & & 4.43 & 0.229 \\
\rowcolor{oursrow} Eco3R ($\pi^3$) & 1.29 & \textbf{0.092} & & \textbf{1.48} & \textbf{0.041} & & \textbf{1.07} & 0.095 & & \textbf{1.44} & \textbf{0.033} \\
  \bottomrule
\end{tabular}
}
\end{table}

\paragraph{Other datasets}
\label{sec:exp_heldout}
On the four other datasets (\cref{tab:heldout}, Sec.~S4), Eco3R ($\pi^3$) stays within 2.4\% of the full $\pi^3$ in accuracy and completeness while running 6.9$\times$ to 10.4$\times$ faster. It reduces the ATE of every chunked and streaming method by at least 20\% on every dataset. Eco3R (VGGT) reduces the accuracy error of VGGT by 7\% to 33\% on every dataset. Compared with the frozen backbone, refocusing reduces this error by 32\% on 7-Scenes and 11\% on TUM RGB-D (Sec.~S6).

\subsection{Ablations and analysis}
\label{sec:exp_ablation}

We conduct ablation studies of Eco3R on ScanNet at 500 frames (\cref{tab:ablation}). Without refocusing, the success rate decreases by 10 points, while AbsRel increases by 29.0\%.
Refocusing increases the correlation between attention and co-visibility in blocks 12--17 (\cref{fig:band}b, ${p < 0.01}$). It works best when applied to the whole band, since tuning only the readout blocks 16--21 or the early blocks 0--5 reduces the success rate by 2 and 6 points, respectively.

\begin{table}[t]
  \caption{\textbf{Ablations of Eco3R} on ScanNet at 500 frames. Shaded rows give absolute values in the units and notation of \cref{tab:main,tab:eff} (Comp.\ in cm). The rows below each shaded row give the changes in success rate (points) and median errors (\%), as well as the ratios of TFLOPs and FPS. Indented rows compare with the Online Stream row. Gray marks changes below 5\% that are not significant (${p \geq 0.05}$).}
  \label{tab:ablation}
  \centering
  \scriptsize
  \setlength{\tabcolsep}{1.6pt}
  \renewcommand{\arraystretch}{0.88}
  \resizebox{\linewidth}{!}{\begin{tabular}{@{}l rrrrrr rr@{}}
  \toprule
  Variant & Succ.$\uparrow$ & ATE$\downarrow$ & RPE$\downarrow$ & Acc.$\downarrow$ & Comp.$\downarrow$ & AbsRel$\downarrow$ & TFLOPs$\downarrow$ & FPS$\uparrow$ \\
  \midrule
\rowcolor{oursrow} Eco3R (VGGT)$^{\dagger}$ & 82 & 0.090 & 0.528 & 1.48 & 1.46 & 0.017 & 5.58 & 21.6 \\
\midrule
w/o Alignment Refocusing & $-$10 & $+$29.7 & $+$8.1 & $+$35.0 & $+$20.9 & $+$29.0 & 1.00 & 1.00 \\
w/o Pinned Key Set & $-$4 & $+$19.8 & $+$7.8 & $+$11.9 & \textcolor{gray}{$+$3.5} & \textcolor{gray}{$+$3.7} & 9.96 & 0.13 \\
w/o both (VGGT) & $-$12 & $+$23.8 & $+$11.4 & $+$32.8 & $+$22.0 & $+$28.8 & 9.96 & 0.12 \\
\midrule
16 pinned frames & $-$4 & $+$13.9 & $+$10.1 & $+$9.9 & $+$3.5 & $+$6.2 & 0.71 & 1.30 \\
63 pinned frames & 0 & \textcolor{gray}{$+$0.8} & $-$3.4 & \textcolor{gray}{$+$0.3} & \textcolor{gray}{$-$1.3} & $-$3.4 & 1.62 & 0.68 \\
MegaLoc keys~\cite{megaloc} & $-$2 & $+$6.3 & \textcolor{gray}{$+$0.1} & \textcolor{gray}{$+$2.7} & \textcolor{gray}{$+$2.5} & \textcolor{gray}{$+$0.1} & 1.02 & 0.99 \\
$+$\,own keys~\cite{kvtracker} & $-$2 & \textcolor{gray}{$+$2.7} & \textcolor{gray}{$-$1.6} & $+$9.3 & \textcolor{gray}{$+$0.4} & \textcolor{gray}{$-$0.8} & — & — \\
$+\,8$ nearby keys & $-$12 & $+$37.5 & $+$2.0 & $+$110.4 & $+$38.9 & $-$2.7 & 1.14 & 0.90 \\
\midrule
Tuning blocks 16--21 & $-$2 & $+$12.7 & $+$9.1 & $+$16.1 & $+$4.0 & $+$14.3 & 1.00 & 1.01 \\
Tuning blocks 0--5 & $-$6 & \textcolor{gray}{$+$3.3} & $+$2.2 & $+$9.2 & $+$6.9 & \textcolor{gray}{$+$1.8} & 1.00 & 0.99 \\
\midrule
Online Stream & $-$4 & $+$13.9 & $+$2.3 & $+$12.6 & \textcolor{gray}{0.0} & $+$1.3 & 1.68 & 0.50 \\
\hspace{0.8em}translation scale & 0 & \textcolor{gray}{$-$1.6} & \textcolor{gray}{0.0} & $+$18.6 & $+$11.5 & \textcolor{gray}{0.0} & 1.00 & 0.98 \\
\hspace{0.8em}w/o rescaling & $-$12 & $+$26.3 & \textcolor{gray}{0.0} & $+$188.7 & $+$97.8 & \textcolor{gray}{0.0} & 1.00 & 1.03 \\
\midrule
\rowcolor{oursrow} Eco3R ($\pi^3$) & 94 & 0.052 & 0.356 & 1.00 & 1.16 & 0.011 & 4.79 & 25.1 \\
\midrule
16 pinned frames & $-$4 & $+$14.7 & $+$7.6 & $+$7.3 & $+$4.3 & $+$9.3 & 0.74 & 1.26 \\
63 pinned frames & 0 & $+$8.0 & $-$2.4 & $-$5.2 & $-$2.1 & $-$4.4 & 1.54 & 0.71 \\
MegaLoc keys~\cite{megaloc} & $-$2 & $+$9.3 & \textcolor{gray}{$-$1.2} & \textcolor{gray}{$+$1.0} & \textcolor{gray}{$+$1.6} & $+$0.6 & 1.02 & 0.97 \\
$+\,8$ nearby keys & $-$2 & $+$41.6 & $+$48.6 & $+$42.2 & $+$6.3 & \textcolor{gray}{$+$0.8} & 1.12 & 0.93 \\
\midrule
\rowcolor{oursrow} Eco3R (VGGT-$\Omega$) & 76 & 0.112 & 0.410 & 1.95 & 2.10 & 0.025 & 2.42 & 42.6 \\
\rowcolor{oursrow} Eco3R (DA3) & 66 & 0.127 & 0.789 & 2.64 & 2.16 & 0.027 & 2.42 & 36.8 \\
  \bottomrule
\end{tabular}
}
\end{table}

On both backbones, reducing the number of pinned frames to 16 lowers the success rate by 4 points and increases RPE by 7.6\%--10.1\%. Increasing it to 63 reduces RPE by 2.4\%--3.4\%, but lowers throughput to approximately 70\% of that achieved with 31 pinned frames.
Using keys selected by content performs worse. For example, using farthest-point sampling on MegaLoc descriptors~\cite{megaloc,gotohunt} to select keys increases the ATE of Eco3R ($\pi^3$) by 9.3\%.
Allowing attached frames to also attend to their own keys~\cite{kvtracker} (+own keys) does not significantly affect ATE or RPE but increases the accuracy error by 9.3\%.
In the Online Stream, estimating  scale from camera translation ratios increases the accuracy error by 18.6\% compared with depth-median-based estimation. Omitting rescaling altogether increases the error by  189\% and reduces the success rate by 12 points. The final rows demonstrate the direct application of Eco3R to VGGT-$\Omega$ and DA3 without retraining, both of which outperform their full models.

\paragraph{Alignment Score}
\label{sec:exp_score}
On the full VGGT, the score achieves an AUROC of 0.91,
compared with 0.61 for the predicted confidence and 0.52 for the normalized attention entropy. Computed in the pin pass of the frozen VGGT, the Alignment Score detects failures with an AUROC of 0.84 at a sixteenth of the cost of scoring all frames, and it agrees with the score computed over all frames (Spearman correlation 0.99).
\section{Conclusion}
\label{sec:conclusion}

Our experiments on long indoor videos show that a small set of uniformly sampled frames, selected independently of video content, can serve as shared keys for every frame. Eco3R reproduces joint inference with these keys exactly at a constant cost per frame, achieving 6.9--16.6$\times$ speedups on VGGT and $\pi^3$ while maintaining or improving success rates. For VGGT, Alignment Refocusing improves success rates, and the training-free Alignment Score detects alignment failures. Eco3R reconstructs up to 4,000 frames on a single A100 GPU and streams at 18.6 fps with the lowest online absolute trajectory error. Feed-forward 3D models can thus scale to long videos through a small set of shared keys.
\section*{LLM Usage Statement}
We used Claude Code (version 2.1.247), an agentic coding assistant built on Claude model (Opus 5.5), in the research and the writing of this paper.
In the research, it helped us automate the evaluation and run the compared methods with their official code.
We did not use any image-generation model.
For the writing, we used it for grammar checking, rephrasing, and formatting.

{
    \small
    \bibliographystyle{ieeenat_fullname}
    \bibliography{main}
}

\clearpage
\setcounter{topnumber}{1}\setcounter{bottomnumber}{1}\setcounter{totalnumber}{2}\setcounter{dbltopnumber}{1}
\renewcommand{\topfraction}{0.8}\renewcommand{\bottomfraction}{0.6}\renewcommand{\textfraction}{0.12}
\renewcommand{\dbltopfraction}{0.8}\renewcommand{\floatpagefraction}{0.85}\renewcommand{\dblfloatpagefraction}{0.7}
\maketitlesupplementary
\setcounter{section}{0}\setcounter{table}{0}\setcounter{figure}{0}\setcounter{algorithm}{0}
\renewcommand{\thesection}{S\arabic{section}}
\renewcommand{\thetable}{S\arabic{table}}
\renewcommand{\thefigure}{S\arabic{figure}}
\renewcommand{\thealgorithm}{S\arabic{algorithm}}
\renewcommand{\theequation}{S\arabic{equation}}
\crefname{algorithm}{Algorithm}{Algorithms}

This supplementary material provides the details and results referenced in the main paper. \Cref{ssec:online} presents the Online Stream as an algorithm and defines the rescaling that maintains a common scale across its passes. \Cref{ssec:methods} describes the protocol and the settings of the compared methods. \Cref{ssec:scannet,ssec:deploy,ssec:heldout,ssec:ablation} report the full results underlying \cref{sec:experiments}. Beyond the ablations, \cref{ssec:ablation} provides the training loss of Alignment Refocusing and the proof of \cref{prop:exact}. \Cref{ssec:analysis} examines whether the readout blocks tolerate fewer query tokens, how nearby keys degrade camera pose accuracy, and how sharper attention affects the full VGGT. \Cref{ssec:score} details the Alignment Score. \Cref{ssec:reloc} reports relocalization on 7-Scenes. \Cref{ssec:videos} shows how three ways of running a multi-view transformer fail on one long video and presents reconstructions of more videos. \Cref{ssec:limits} discusses the limitations and future work, including trajectories on KITTI odometry. As in the main paper, Eco3R ($X$) denotes Eco3R on backbone $X$. As in the main paper, we report scene-level medians, with all relative changes calculated from these median values. We assess significance with Wilcoxon signed-rank tests over scenes. The trajectory error refers to the absolute trajectory error (ATE) defined in \cref{ssec:scannet}.

\section{Online Stream}
\label{ssec:online}

\begin{figure}[t]
  \centering
  \includegraphics[width=\linewidth]{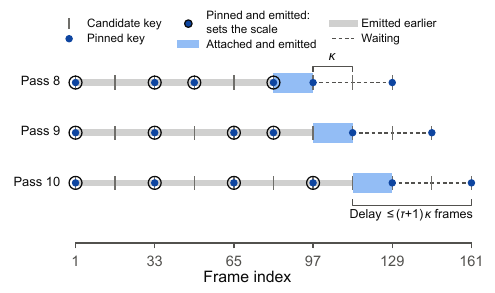}
  \caption{\textbf{Online Stream.} Three consecutive passes for the key interval ${\kappa = 16}$, the key budget ${M_{\max} = 6}$, and the lag ${\tau = 2}$. Each pass runs a new pin pass over at most $M_{\max}$ candidates uniformly sampled from the past. It then attaches and emits the $\kappa$ frames up to the candidate that arrived $\tau$ passes earlier.}
  \label{sfig:stream}
\end{figure}

\Cref{salg:online} details the Online Stream of \cref{sec:method_online}, while \cref{sfig:stream} illustrates three consecutive passes. Every $\kappa$ frames, the newest frame becomes a candidate key. Each candidate closes the key interval of the $\kappa$ frames up to it and starts a new pass. Pass $m$ first runs a pin pass over the key set $\mathcal{P}_m$ of \cref{eq:thin} and stores the resulting Scene Cache. The pass then attaches the frames of the key interval that closed $\tau$ passes earlier and rescales their outputs to the scale of the stream. Let $D_k^{(m)}$ denote the depth map that pass $m$ predicts for frame $k$, and let $\hat{D}_k$ denote the emitted depth map of frame $k$. Let $\mathcal{S}_m \subseteq \mathcal{P}_m$ denote the pinned frames of pass $m$ that earlier passes have emitted. For ${m > 0}$, we rescale the outputs of pass $m$ as
\begin{equation}
  \begin{gathered}
    s_m = \operatorname*{median}_{k \in \mathcal{S}_m} \frac{\mathrm{med}\big(\hat{D}_k\big)}{\mathrm{med}\big(D_k^{(m)}\big)}, \\
    \hat{D}_i = s_m D_i^{(m)}, \qquad \hat{\mathbf{t}}_i = s_m \mathbf{t}_i^{(m)},
  \end{gathered}
  \label{eq:stitch}
\end{equation}
for every frame $i$ that pass $m$ emits, where $\mathrm{med}$ denotes the median over pixels, $\mathbf{t}_i^{(m)}$ the camera translation that pass $m$ predicts for frame $i$, and $\hat{\mathbf{t}}_i$ the emitted camera translation of frame $i$. We set ${s_0 = 1}$ for the first pass and clip $s_m$ to $[1/3, 3]$ for later passes. For $\pi^3$, which has no reference frame, a rigid transform~\cite{umeyama} then maps the rescaled cameras of the frames in $\mathcal{S}_m$ onto their emitted poses. When the video ends, a final pass emits the remaining frames. Let $t_{\mathrm{pin}}$ and $t_{\mathrm{att}}$ denote the time of one pin pass and the time to attach one frame, respectively. Processing a video of $N$ frames then takes approximately ${(N/\kappa)\, t_{\mathrm{pin}} + N t_{\mathrm{att}}}$. The cost per frame remains constant once the key set reaches the key budget $M_{\max}$.

\begin{algorithm}[t]
  \caption{Online Stream}
  \label{salg:online}
  \begin{algorithmic}[1]
    \Require frames arriving in order, key interval $\kappa$, key budget $M_{\max}$, and lag $\tau$
    \Ensure pose and depth of every frame, emitted at most $(\tau+1)\kappa$ frames after its arrival, and the Alignment Score $\alpha_m$ of every pass
    \State $m \gets 0$, $\ \mathcal{E} \gets \emptyset$ \Comment{$\mathcal{E}$: emitted frames}
    \For{each arriving frame $i$}
      \If{$i = 1 + m\kappa$} \Comment{$i$ is the candidate $c_m$}
        \State $\mathcal{P}_m \gets$ key set of \cref{eq:thin}
        \State $\mathcal{C}, \alpha_m \gets$ pin pass on $\mathcal{P}_m$ \Comment{\cref{sec:method_cache}}
        \State $\mathcal{A} \gets \{\, j \le c_{\max(m-\tau,\, 0)} \,\} \setminus \mathcal{E}$ \Comment{frames to emit}
        \For{each batch $\mathcal{B} \subseteq \mathcal{A} \setminus \mathcal{P}_m$}
          \State attach pass on $\mathcal{B}$ with $\mathcal{C}$ \Comment{\cref{eq:pass}}
        \EndFor
        \State $s_m \gets$ \cref{eq:stitch} on $\mathcal{S}_m = \mathcal{P}_m \cap \mathcal{E}$
        \State for $\pi^3$, fit the rigid transform on $\mathcal{S}_m$
        \State emit the frames of $\mathcal{A}$ scaled by $s_m$, and $\mathcal{E} \gets \mathcal{E} \cup \mathcal{A}$
        \State $m \gets m + 1$
      \EndIf
    \EndFor
    \State at the end of the video, a final pass over the last key set emits the remaining frames
  \end{algorithmic}
\end{algorithm}

We first selected the key interval $\kappa$ and the lag $\tau$ using three evaluation scenes. To verify these settings, we repeated the comparison between the online and offline results of Eco3R (VGGT) on the first 1,000 consecutive frames of ten ScanNet training scans. With ${\tau = 2}$, setting ${\kappa = 16}$ or ${\kappa = 32}$ maintains the trajectory error within 1.1\% and the relative rotation error within 4.6\% of the offline results. The root mean square angle between the aligned and ground-truth camera orientations exceeds 10\textdegree{} online only in the scenes where it exceeds 10\textdegree{} offline. Running without lag at ${\kappa = 16}$ increases the relative rotation error by 7.1\% relative to the offline result. Setting ${\kappa = 64}$ with ${\tau = 2}$ increases the trajectory error by 9.3\% relative to the offline result. These results on the training scans verify the settings of the main paper. For $\pi^3$, the Online Stream uses the same key interval and key budget as for VGGT and runs without lag.

\section{Compared methods}
\label{ssec:methods}

\paragraph{Protocol}
All methods process the frames selected by the protocol of FastVGGT~\cite{fastvggt} (\cref{ssec:scannet}) at a resolution of 518 $\times$ 392 pixels, unless noted otherwise below. Each compared method runs its official code with its released weights. We exclude KV-Tracker~\cite{kvtracker}, which predicts only camera poses for frames outside its keyframes. In all tables, $^{\dagger}$ marks the methods that train or fine-tune network weights. All other methods, apart from the full models, are training-free. Eco3R attaches the frames outside the Pinned Key Set in batches of 64 frames. For offline Eco3R, the camera head of VGGT~\cite{vggt} processes all frames. Unless noted otherwise, each run uses one 80\,GB A100 GPU with 8--12 CPU cores and 96--240\,GB of host memory. The key-count experiments on training scans (\cref{ssec:ablation}), the analyses of nearby keys (\cref{ssec:analysis}), and the VGGT-$\Omega$~\cite{vggtomega} experiments on the other datasets ran on a 24\,GB workstation GPU. SAIL-Recon~\cite{sailrecon} and DA-VGGT~\cite{divpart} ran on this workstation GPU for the accuracy evaluation and on the A100 GPU for the time and memory measurements. By default, each method stores its weights in single precision and computes under bfloat16 autocast, as done in its official inference code.

\paragraph{Time and memory}
We measure the runtime of each scene between CUDA synchronizations before and after inference. The runtime includes moving the frames to the GPU and converting the outputs for our evaluation, but excludes decoding and resizing the images. For VGGT-Long~\cite{vggtlong}, VGGT-SLAM~\cite{vggtslam}, CUT3R~\cite{cut3r}, TTT3R~\cite{ttt3r}, Good Token Hunting~\cite{gotohunt}, SAIL-Recon, and DA-VGGT, we use the runtimes reported by their code. The peak memory is defined as the largest amount of GPU memory that PyTorch allocates during inference. It includes the weights but excludes the input frames, which a method can store in host memory and move to the GPU as needed. \Cref{tab:scale} counts each scene that exceeds 80\,GB of memory as a failure. The full VGGT and the full $\pi^3$ exceed 80\,GB of memory on scenes longer than 1,755 and 1,812 frames, respectively. Following the official code of $\pi^3$, we store a single copy of the input frames on the GPU in both runs. At 500 and 1,000 frames, we rerun the scenes that exceed 80\,GB of memory with CUDA unified memory. We include the accuracy of these reruns in the results but omit their time and memory. As StreamVGGT~\cite{streamvggt} exceeds 80\,GB of memory at 1,000 frames in all but one of the tested scenes, we do not rerun it and list it as out of memory in the tables. On scenes that fit in GPU memory, unified memory reproduces the regular runs bit for bit for FlashVGGT~\cite{flashvggt}, Depth Anything 3 (DA3)~\cite{da3}, Good Token Hunting, and the block-sparse attention of Wang et al.~\cite{sparsevggt}. For VGG-T$^3$~\cite{vggttt} and SAF3R~\cite{saf3r}, the unified-memory runs agree with the regular runs to within 0.04\% and 0.4\%, respectively. The difference for SAF3R may arise from its Triton kernels, which tune differently under unified memory. We measure latency and energy in a first job that runs the methods in turn on a single A100 GPU. Later jobs on the same type of GPU add CUT3R, TTT3R, RetrieveVGGT~\cite{retrievevggt}, and VGGT-$\Omega$ to both measurements. For VGG-T$^3$, these later jobs measure only the energy. In each later energy job, the energy of offline Eco3R ($\pi^3$) and LASER~\cite{laser} remains within 1.1\% of that measured in the first job.

\paragraph{Offline methods}
VGGT and FastVGGT run with the VGGT-1B weights in the evaluation code of FastVGGT, which casts the weights and the prediction heads to bfloat16. Eco3R (VGGT) runs in the same precision. Following its official command, FastVGGT merges tokens from global block 0 with a merging ratio of 0.9. DA3 normalizes the frames with the ImageNet statistics of its interface. Like the other methods, DA3 expresses its predictions in the coordinate frame of the first camera. As in its README and evaluation configuration, FlashVGGT downsamples the keys and values of each frame by a factor of 4 and selects one keyframe per 200 frames by k-means clustering of the frame tokens. FlashVGGTStream, the streaming variant of FlashVGGT, uses the setting of its official demo. FlashVGGTStream processes chunks of 10 frames and downsamples the keys and values by a factor of 3, which is the default in its code. As its camera head predicts the poses only after the last chunk, we group FlashVGGTStream with the offline methods. VGG-T$^3$ runs in the memory-efficient mode of its evaluation configuration. Following its interface, VGG-T$^3$ adapts the number of test-time training steps to the video length, using ${2 + \lfloor N/1000 \rfloor}$ steps for $N$ frames. SAIL-Recon runs the inference of its demo script with its released weights. It samples 100 anchor frames uniformly in time and stores 300 randomly selected tokens per anchor frame in the key-value cache of each block. SAIL-Recon then processes batches of 20 frames, each of which attends only to this cache and its own tokens. DA-VGGT runs the training-free pipeline of its 3D evaluation script on VGGT. This pipeline partitions the frames into chunks of 50 frames based on the dissimilarity of their DINOv2 features~\cite{dinov2} and the spread of the camera positions estimated from a seed chunk. The pipeline then processes each chunk with the first frame as a shared anchor and aligns the chunks through the pose and depth of this anchor. LiteVGGT~\cite{litevggt} runs on an H100 GPU with its bfloat16 weights and its aggregator in 8-bit floating point (FP8), as in its own evaluation (\cref{ssec:scannet}). We apply the block-sparse attention of Wang et al.~\cite{sparsevggt} to every global block with a sparse ratio of 0.6 and a cumulative attention threshold of 0.4. This attention runs on the 8-bit integer and 16-bit floating-point attention kernels of its authors. SAF3R uses the per-head patterns of its repository with four keys per frame for its top-$k$ selection. Good Token Hunting restricts the keys to 25 frames that farthest-point sampling selects on MegaLoc descriptors~\cite{megaloc}. It replaces global blocks 0 and 1 with frame attention and subsamples the key tokens of these frames with a stride of 3 along both image axes in global blocks 2--8. On VGGT, Good Token Hunting runs in bfloat16, like VGGT.

\paragraph{Chunked pipelines}
VGGT-Long runs the configuration of its repository. It processes chunks of 60 frames that overlap by 30 frames and detects loop closures with SALAD descriptors~\cite{salad}, using a similarity threshold of 0.85 and the top 5 candidates. It aligns the chunks by similarity transforms estimated with iteratively reweighted least squares. The loop optimization of VGGT-Long uses the Python implementation of its repository. The runtime of VGGT-Long covers the whole pipeline. As in its demo script, LASER runs without its optional loop closure. It processes windows of 30 frames that overlap by 10 frames, with depth refinement and a top-confidence percentile of 0.3. VGGT-SLAM runs version 1.0 of its official code with submaps of 32 frames (the setting of its main results). It uses one overlapping frame between submaps, at most one loop closure per submap, and its default optimization on the SL(4) manifold of projective transformations. The keyframe gate of VGGT-SLAM, which thresholds the optical flow between frames, retains only 8\%--76\% of the frames of our videos and produces no output for the other frames. To obtain an output for every frame, we enable the option in its code that treats every frame as a keyframe. We run VGGT-SLAM once with seed 0.

\paragraph{Streaming methods}
StreamVGGT attends causally to a cache of all past frames. As in its reconstruction script, RetrieveVGGT uses the first frame as an anchor and retrieves 47 past frames by segment sampling. Its segment sampling uses a threshold at the mean query-key similarity plus 0.3 standard deviations. InfiniteVGGT~\cite{infinitevggt} uses a budget of 1.2 million tokens. FrameVGGT~\cite{framevggt} runs its official long-sequence mode with the weights of StreamVGGT. We count the runs that overflow its buffer of 32,768 tokens per layer as failures. As done in their official code for a new sequence, we reset the key-value store of RetrieveVGGT and the state of InfiniteVGGT for every scene. CUT3R and TTT3R share the weights of CUT3R and differ only in their state update rules. We run both methods with the official code of TTT3R, which implements both update rules. Following the evaluation of TTT3R, both methods take inputs of 512 $\times$ 368 pixels and run in the single precision of its official code. On 16 frames, the TTT3R code reproduces the points predicted by the official CUT3R code with a median relative difference of $2 \times 10^{-4}$. MUSt3R~\cite{must3r} runs its 512-pixel model in single precision on inputs of 512 $\times$ 384 pixels. Offline, as in the default setting of its export script, MUSt3R builds its memory from 50 frames uniformly sampled over the video. It then predicts every frame from this memory. Online, MUSt3R runs the slam-keyframe mode of its demo with the default settings. In this mode, each frame attends to the memory of the earlier frames, while new keyframes extend this memory. We map the predicted depth, points, and focal length of CUT3R, TTT3R, and MUSt3R onto the 518 $\times$ 392 pixels of our evaluation and leave the cropped pixels without predictions. On KITTI, LongStream~\cite{longstream} runs in its streaming refresh mode with a keyframe stride of 8 and a refresh every 4 keyframes. We evaluate the poses of LongStream with our evaluation code.

\section{Full results on ScanNet}
\label{ssec:scannet}

\begin{table*}[p]
  \caption{\textbf{ScanNet at 500 and 1,000 frames}, extending \cref{tab:main}. Units are percent for success rate (Succ.), meters for absolute trajectory error (ATE), degrees and centimeters for relative rotation and translation errors (RPE-r and RPE-t), centimeters for accuracy (Acc.) and completeness (Comp.), seconds per scene for runtime, and GB for peak inference memory (maximum across scenes, \cref{ssec:methods}). The absolute relative depth error (AbsRel) has no unit. The first three rows at each length are the full models. Other notation follows \cref{tab:main}, except that ``failed'' marks a success rate below 10\%. Dashes mark the values of failed rows and memory that we did not measure. $^{\ddagger}$Time of the full pipeline, including inference times of 74\,s and 151\,s at 500 and 1,000 frames, respectively. $^{\|}$Run with unified memory in the 84\% of scenes that exceed 80\,GB of memory at 1,000 frames, with time and memory omitted. FrameVGGT stops early on 11 and 24 scenes and VGGT-SLAM on 1 and 6 scenes at 500 and 1,000 frames, respectively. These scenes count as failures (\cref{ssec:methods}).}
  \label{stab:scannet}
  \centering
  \footnotesize
  \renewcommand{\arraystretch}{0.86}
  \begin{tabular}{@{}l rrrrrrrrr@{}}
  \toprule
  Method & Succ.$\uparrow$ & ATE$\downarrow$ & RPE-r$\downarrow$ & RPE-t$\downarrow$ & AbsRel$\downarrow$ & Acc.$\downarrow$ & Comp.$\downarrow$ & Time$\downarrow$ & Mem.$\downarrow$ \\
  \midrule
\multicolumn{10}{c}{500 frames} \\
  \midrule
VGGT~\cite{vggt} & 70 & 0.111 & 0.589 & 2.59 & 0.0218 & 1.97 & 1.78 & 196 & 23.1 \\
$\pi^3$~\cite{pi3} & 94 & 0.054 & 0.343 & 1.47 & 0.0109 & 0.94 & 1.15 & 138 & 24.4 \\
DA3~\cite{da3} & 64 & 0.137 & 1.004 & 3.62 & 0.0269 & 2.61 & 2.15 & 66.1 & — \\
\midrule
\multicolumn{10}{@{}l}{\textit{Offline}} \\
FastVGGT~\cite{fastvggt} & 74 & 0.098 & 0.598 & 2.58 & 0.0222 & 1.95 & 1.73 & 68.8 & 32.3 \\
FlashVGGT~\cite{flashvggt}$^{\dagger}$ & 76 & 0.121 & 0.563 & 2.05 & 0.0177 & 1.88 & 1.76 & 19.7 & 38.0 \\
FlashVGGTStream~\cite{flashvggt}$^{\dagger}$ & 26 & 0.307 & 2.013 & 6.67 & 0.0237 & 8.70 & 4.86 & 20.5 & 18.0 \\
VGG-T$^3$~\cite{vggttt}$^{\dagger}$ & 14 & 0.456 & 5.479 & 10.75 & 0.0303 & 8.08 & 3.65 & 32.5 & — \\
SAIL-Recon~\cite{sailrecon}$^{\dagger}$ & 70 & 0.118 & 0.492 & 2.29 & 0.0216 & 1.95 & 1.92 & 104 & 13.1 \\
DA-VGGT~\cite{divpart} & 72 & 0.118 & 0.671 & 3.66 & 0.0216 & 2.03 & 1.56 & 38.7 & 13.1 \\
MUSt3R~\cite{must3r}$^{\dagger}$ & 70 & 0.130 & 0.534 & 2.84 & 0.0271 & 2.62 & 2.33 & 59.8 & \textbf{4.6} \\
VGGT-Long~\cite{vggtlong}$^{\ddagger}$ & 68 & 0.129 & 0.376 & 1.67 & 0.0224 & 2.18 & 2.43 & 263 & 20.4 \\
\rowcolor{oursrow} Eco3R (VGGT)$^{\dagger}$ & 82 & 0.090 & 0.528 & 1.90 & 0.0170 & 1.48 & 1.46 & 23.1 & 10.0 \\
\rowcolor{oursrow} Eco3R ($\pi^3$) & \textbf{94} & \textbf{0.052} & \textbf{0.356} & \textbf{1.55} & \textbf{0.0115} & \textbf{1.00} & \textbf{1.16} & 19.9 & 9.9 \\
\rowcolor{oursrow} Eco3R (DA3) & 66 & 0.127 & 0.789 & 3.05 & 0.0269 & 2.64 & 2.16 & \textbf{13.6} & — \\
\midrule
\multicolumn{10}{@{}l}{\textit{Online}} \\
CUT3R~\cite{cut3r}$^{\dagger}$ & failed & — & — & — & — & — & — & — & — \\
TTT3R~\cite{ttt3r} & 20 & 0.352 & 0.751 & 3.30 & 0.0280 & 7.86 & 3.86 & \textbf{27.5} & \textbf{6.1} \\
MUSt3R~\cite{must3r}$^{\dagger}$ & 64 & 0.148 & 0.620 & 3.23 & 0.0303 & 3.07 & 2.12 & 69.3 & 9.6 \\
StreamVGGT~\cite{streamvggt}$^{\dagger}$ & 48 & 0.210 & 1.191 & 4.13 & 0.0234 & 6.20 & 3.67 & 456 & 64.9 \\
RetrieveVGGT~\cite{retrievevggt} & 38 & 0.235 & 2.534 & 7.32 & 0.0244 & 7.26 & 3.43 & 407 & 12.6 \\
InfiniteVGGT~\cite{infinitevggt} & 30 & 0.262 & 2.627 & 5.97 & 0.0253 & 6.83 & 3.64 & 127 & 14.9 \\
FrameVGGT~\cite{framevggt} & 22 & 0.277 & 4.559 & 5.59 & 0.0238 & 6.92 & 3.16 & 871 & 20.4 \\
LASER~\cite{laser} & 84 & 0.107 & \textbf{0.316} & \textbf{1.24} & 0.0139 & 2.05 & 1.74 & 43.9 & 6.6 \\
VGGT-SLAM~\cite{vggtslam} & 50 & 0.192 & 0.468 & 1.91 & 0.0334 & 3.81 & 3.54 & 41.9 & 10.8 \\
\rowcolor{oursrow} Eco3R (VGGT)$^{\dagger}$ & 78 & 0.102 & 0.540 & 1.94 & 0.0172 & 1.67 & 1.46 & 45.8 & 11.4 \\
\rowcolor{oursrow} Eco3R ($\pi^3$) & \textbf{90} & \textbf{0.061} & 0.368 & 1.55 & \textbf{0.0124} & \textbf{1.22} & \textbf{1.24} & 37.0 & 11.0 \\
  \midrule[\heavyrulewidth]
\multicolumn{10}{c}{1,000 frames} \\
  \midrule
VGGT~\cite{vggt} & 70 & 0.130 & 0.581 & 2.47 & 0.0230 & 2.26 & 1.73 & 772 & 44.2 \\
$\pi^3$~\cite{pi3} & 94 & 0.058 & 0.265 & 1.04 & 0.0110 & 0.90 & 1.13 & 557 & 43.4 \\
DA3~\cite{da3}$^{\|}$ & 70 & 0.168 & 0.950 & 3.41 & 0.0275 & 2.61 & 2.05 & — & — \\
\midrule
\multicolumn{10}{@{}l}{\textit{Offline}} \\
FastVGGT~\cite{fastvggt} & 72 & 0.100 & 0.521 & 2.09 & 0.0221 & 2.00 & 1.75 & 196 & 62.5 \\
FlashVGGT~\cite{flashvggt}$^{\dagger}$ & 76 & 0.118 & 0.426 & 1.65 & 0.0178 & 1.80 & 1.70 & 50.6 & 70.1 \\
FlashVGGTStream~\cite{flashvggt}$^{\dagger}$ & 18 & 0.337 & 3.452 & 5.67 & 0.0246 & 9.76 & 4.28 & 50.0 & 29.2 \\
VGG-T$^3$~\cite{vggttt}$^{\dagger}$ & 10 & 0.354 & 1.541 & 6.15 & 0.0287 & 6.16 & 3.36 & 72.5 & — \\
SAIL-Recon~\cite{sailrecon}$^{\dagger}$ & 72 & 0.108 & 0.405 & 1.76 & 0.0215 & 2.01 & 1.89 & 199 & 13.1 \\
DA-VGGT~\cite{divpart} & 72 & 0.119 & 0.580 & 2.95 & 0.0217 & 2.08 & 1.54 & 78.0 & 13.5 \\
MUSt3R~\cite{must3r}$^{\dagger}$ & 70 & 0.133 & 0.445 & 2.44 & 0.0269 & 2.53 & 2.23 & 117 & \textbf{5.7} \\
VGGT-Long~\cite{vggtlong}$^{\ddagger}$ & 62 & 0.164 & 0.328 & 1.44 & 0.0236 & 2.70 & 2.63 & 547 & 20.5 \\
\rowcolor{oursrow} Eco3R (VGGT)$^{\dagger}$ & 82 & 0.093 & 0.434 & 1.43 & 0.0173 & 1.53 & 1.52 & 46.5 & 10.0 \\
\rowcolor{oursrow} Eco3R ($\pi^3$) & \textbf{94} & \textbf{0.060} & \textbf{0.281} & \textbf{1.16} & \textbf{0.0113} & \textbf{1.00} & \textbf{1.16} & 40.4 & 9.9 \\
\rowcolor{oursrow} Eco3R (DA3) & 70 & 0.128 & 0.799 & 2.89 & 0.0271 & 2.53 & 2.15 & \textbf{27.4} & — \\
\midrule
\multicolumn{10}{@{}l}{\textit{Online}} \\
CUT3R~\cite{cut3r}$^{\dagger}$ & failed & — & — & — & — & — & — & — & — \\
TTT3R~\cite{ttt3r} & failed & — & — & — & — & — & — & — & — \\
MUSt3R~\cite{must3r}$^{\dagger}$ & 68 & 0.149 & 0.499 & 2.77 & 0.0300 & 3.06 & 2.10 & 116 & 11.3 \\
StreamVGGT~\cite{streamvggt}$^{\dagger}$ & \multicolumn{9}{c}{out of memory} \\
RetrieveVGGT~\cite{retrievevggt} & 38 & 0.281 & 2.552 & 6.91 & 0.0245 & 7.64 & 3.35 & 856 & 12.7 \\
InfiniteVGGT~\cite{infinitevggt} & 20 & 0.361 & 5.666 & 6.02 & 0.0282 & 11.92 & 5.34 & 263 & 15.0 \\
FrameVGGT~\cite{framevggt} & 20 & 0.232 & 3.607 & 3.27 & 0.0227 & 6.99 & 2.75 & 1726 & 20.0 \\
LASER~\cite{laser} & 80 & 0.120 & \textbf{0.232} & \textbf{0.91} & 0.0150 & 2.61 & 2.25 & 87.0 & \textbf{6.6} \\
VGGT-SLAM~\cite{vggtslam} & 48 & 0.183 & 0.448 & 1.88 & 0.0362 & 5.26 & 4.73 & 90.4 & 12.1 \\
\rowcolor{oursrow} Eco3R (VGGT)$^{\dagger}$ & 78 & 0.099 & 0.449 & 1.52 & 0.0173 & 1.69 & 1.45 & 64.6 & 11.5 \\
\rowcolor{oursrow} Eco3R ($\pi^3$) & \textbf{92} & \textbf{0.061} & 0.280 & 1.14 & \textbf{0.0125} & \textbf{1.13} & \textbf{1.20} & \textbf{53.8} & 11.0 \\
  \bottomrule
\end{tabular}
\end{table*}

\paragraph{Metrics}
The protocol of FastVGGT~\cite{fastvggt} starts from the first frame of each video and samples frames at the largest integer stride that fills the frame budget. At budgets of 500, 1,000, 2,000, and 4,000 frames, the protocol selects a contiguous prefix of the frames for 10, 27, 9, and 3 videos, respectively. At the same budgets, 1, 11, 38, and 47 videos, respectively, are shorter than the budget and retain all their frames. We compute the trajectory metrics using the evaluation code of FastVGGT. This code aligns the predicted trajectory to the ground truth with a similarity transform and then maps the first predicted pose onto the first ground-truth pose. The absolute trajectory error (ATE) is the root mean square of the translation errors of the world-to-camera transforms after this alignment. The relative rotation and translation errors (RPE-r and RPE-t) are the root mean squares of the rotation and translation errors between consecutive frames, respectively. For the point clouds of ScanNet, we back-project every pixel in the central 224 $\times$ 224 region of each frame to a ground-truth point using the ground-truth depth and pose. We pair each ground-truth point with the predicted point of the same pixel and retain at most 999,999 pairs. We align the predicted points to the ground truth with a similarity transform fitted to these pairs and then refine the alignment by point-to-point Iterative Closest Point (ICP) registration with a 0.1\,m threshold. The accuracy (Acc.) and completeness (Comp.) are the mean nearest-neighbor distances from the predicted to the ground-truth points and from the ground-truth to the predicted points, respectively. To compute the absolute relative depth error (AbsRel), we first scale each predicted depth map by its median ratio to the ground truth.

\paragraph{Results at 500 and 1,000 frames}
\Cref{stab:scannet} extends \cref{tab:main} with RPE-t, AbsRel, and Comp., as well as with rows for DA3, Eco3R (DA3), FrameVGGT, and offline MUSt3R. \Cref{stab:sparse} lists Good Token Hunting together with the sparse attention methods. At 500 frames, Eco3R (DA3) succeeds in 66\% of the scenes, compared with 64\% for the full DA3 model, while achieving a 4.9$\times$ speedup. At 1,000 frames, the full DA3 exceeds 80\,GB of memory in 84\% of the scenes. We run these scenes with CUDA unified memory. The full DA3 and Eco3R (DA3) both succeed in 70\% of the scenes. On the 16\% of scenes that fit in 80\,GB of memory, Eco3R (DA3) achieves a 7.0$\times$ speedup over the full DA3. At both frame lengths, Eco3R (DA3) preserves the trajectory accuracy of the full DA3 model. This suggests that the Pinned Key Set transfers to DA3.

\begin{table*}[t]
  \caption{\textbf{Sparse attention and key selection on ScanNet.} RPE is the relative rotation error (RPE-r in \cref{stab:scannet}). Units are as in \cref{stab:scannet}. The first two rows are the full models. Shaded rows are Eco3R, while bold values denote the best results in each column of each backbone group. $^{\|}$Run with unified memory in the 78\% of scenes that exceed 80\,GB of memory at 1,000 frames, with time and memory omitted.}
  \label{stab:sparse}
  \centering
  \footnotesize
  \setlength{\tabcolsep}{3.7pt}
  \begin{tabular}{@{}l rrrrrr c rrrrrr@{}}
  \toprule
  & \multicolumn{6}{c}{500 frames} & & \multicolumn{6}{c}{1,000 frames} \\
  \cmidrule(lr){2-7} \cmidrule(lr){9-14}
  Method & Succ.$\uparrow$ & ATE$\downarrow$ & RPE$\downarrow$ & Acc.$\downarrow$ & Time$\downarrow$ & Mem.$\downarrow$ & & Succ.$\uparrow$ & ATE$\downarrow$ & RPE$\downarrow$ & Acc.$\downarrow$ & Time$\downarrow$ & Mem.$\downarrow$ \\
  \midrule
VGGT~\cite{vggt} & 70 & 0.111 & 0.589 & 1.97 & 196 & 23.1 &  & 70 & 0.130 & 0.581 & 2.26 & 772 & 44.2 \\
$\pi^3$~\cite{pi3} & 94 & 0.054 & 0.343 & 0.94 & 138 & 24.4 &  & 94 & 0.058 & 0.265 & 0.90 & 557 & 43.4 \\
\midrule
\multicolumn{14}{@{}l}{\textit{VGGT backbone}} \\
Wang et al.~\cite{sparsevggt}$^{\|}$ & 62 & 0.113 & 0.673 & 2.43 & 85.4 & 41.2 &  & 62 & 0.132 & 0.678 & 2.53 & — & — \\
SAF3R~\cite{saf3r}$^{\|}$ & 74 & 0.110 & 0.543 & 1.94 & 56.7 & 39.9 &  & 72 & 0.114 & 0.469 & 2.07 & — & — \\
Good Token Hunting~\cite{gotohunt} & 70 & 0.121 & 0.554 & 2.05 & 36.7 & 34.7 &  & 72 & 0.121 & 0.462 & 2.12 & 73.7 & 66.2 \\
\rowcolor{oursrow} Eco3R (VGGT)$^{\dagger}$ & \textbf{82} & \textbf{0.090} & \textbf{0.528} & \textbf{1.48} & \textbf{23.1} & \textbf{10.0} &  & \textbf{82} & \textbf{0.093} & \textbf{0.434} & \textbf{1.53} & \textbf{46.5} & \textbf{10.0} \\
\midrule
\multicolumn{14}{@{}l}{\textit{$\pi^3$ backbone}} \\
Wang et al.~\cite{sparsevggt} & 84 & 0.083 & 0.383 & 1.87 & 64.6 & 29.4 &  & 86 & 0.092 & 0.300 & 1.83 & 256 & 68.2 \\
SAF3R~\cite{saf3r} & 90 & 0.060 & \textbf{0.346} & \textbf{1.00} & 32.1 & 28.2 &  & 90 & 0.063 & \textbf{0.269} & \textbf{0.95} & 101 & 61.4 \\
Good Token Hunting~\cite{gotohunt} & 90 & 0.062 & 0.358 & 1.06 & 23.0 & 25.2 &  & 92 & 0.067 & 0.280 & 1.03 & 45.4 & 44.2 \\
\rowcolor{oursrow} Eco3R ($\pi^3$) & \textbf{94} & \textbf{0.052} & 0.356 & \textbf{1.00} & \textbf{19.9} & \textbf{9.9} &  & \textbf{94} & \textbf{0.060} & 0.281 & 1.00 & \textbf{40.4} & \textbf{9.9} \\
  \bottomrule
\end{tabular}
\end{table*}

\paragraph{Streaming methods}
In the online setting at 500 and 1,000 frames, both versions of Eco3R reduce the median ATE of StreamVGGT, RetrieveVGGT, and InfiniteVGGT by 51\%--83\% (\cref{stab:scannet}), achieving a lower ATE than each of these methods in at least 48 of the 50 scenes (${p < 10^{-12}}$).

\paragraph{VGGT-SLAM}
At 500 and 1,000 frames, VGGT-SLAM succeeds in 50\% and 48\% of the scenes, respectively, compared with 78\% for online Eco3R (VGGT) at both frame lengths and 90\% and 92\% for online Eco3R ($\pi^3$). VGGT-SLAM's median trajectory errors of 0.192\,m and 0.183\,m remain higher than those of both online versions of Eco3R.

\paragraph{MUSt3R}
MUSt3R succeeds in 70\% of the scenes offline, while online it succeeds in 64\% and 68\% at 500 and 1,000 frames, respectively. Offline and online, Eco3R ($\pi^3$) succeeds in at least 90\% of the scenes. In both settings, Eco3R ($\pi^3$) reduces MUSt3R's trajectory and accuracy errors by 55\%--63\% while running 1.9$\times$ to 3.0$\times$ faster.

\paragraph{Sparse attention and key selection}
SAF3R~\cite{saf3r} and the block-sparse attention of Wang et al.~\cite{sparsevggt} skip weak interactions in global attention, while Good Token Hunting~\cite{gotohunt} restricts the keys to frames selected by farthest-point sampling on MegaLoc descriptors~\cite{megaloc}. We run each method on both VGGT and $\pi^3$ with the settings described in \cref{ssec:methods} (\cref{stab:sparse}). On both backbones and at both frame lengths, Eco3R (VGGT) and Eco3R ($\pi^3$) succeed more often than each of the three methods and achieve a significantly lower trajectory error while running 1.1$\times$ to 6.3$\times$ faster.

\paragraph{LiteVGGT}
LiteVGGT~\cite{litevggt} fine-tunes VGGT, merges tokens, and runs its matrix multiplications in FP8. As A100 GPUs do not support FP8, we compare LiteVGGT with the full VGGT, FastVGGT, and Eco3R (VGGT) on one H100 GPU under our protocol (\cref{stab:lite}). Since the FP8 kernels of LiteVGGT require a number of tokens divisible by eight, all four methods use a budget of 496 frames in place of 500. Videos shorter than the budget discard up to seven frames at their end. At 496 and 1,000 frames, Eco3R (VGGT) succeeds in 82\% and 80\% of the scenes, respectively, compared with 68\% and 64\% for LiteVGGT, while running 1.1$\times$ and 1.8$\times$ faster. Eco3R (VGGT) reduces LiteVGGT's ATE by 26\% and 37\%, RPE-r and RPE-t by 21\%--50\%, and AbsRel by 30\% and 29\%, with lower errors in 84\%--100\% of the scenes. In this comparison, LiteVGGT has higher pose and depth errors than the full VGGT. All of these differences are significant except for RPE-r at 1,000 frames.

\begin{table}[H]
  \caption{\textbf{LiteVGGT on one H100 GPU.} Units are as in \cref{stab:scannet}. Shaded rows are Eco3R, while bold values denote the best results at each frame length.}
  \label{stab:lite}
  \centering
  \footnotesize
  \setlength{\tabcolsep}{3pt}
  \resizebox{\linewidth}{!}{\begin{tabular}{@{}l rrrrrrr@{}}
  \toprule
  Method & Succ. & ATE & RPE-r & RPE-t & AbsRel & Acc. & Time \\
  \midrule
\multicolumn{8}{@{}l}{\textit{496 frames}} \\
VGGT~\cite{vggt} & 68 & 0.111 & 0.590 & 2.57 & 0.0219 & 1.99 & 92.2 \\
FastVGGT~\cite{fastvggt} & 74 & 0.099 & 0.597 & 2.58 & 0.0223 & 1.96 & 24.1 \\
LiteVGGT~\cite{litevggt}$^{\dagger}$ & 68 & 0.127 & 0.677 & 3.09 & 0.0235 & 2.31 & 13.0 \\
\rowcolor{oursrow} Eco3R (VGGT)$^{\dagger}$ & \textbf{82} & \textbf{0.094} & \textbf{0.537} & \textbf{1.86} & \textbf{0.0165} & \textbf{1.48} & \textbf{11.8} \\
\midrule
\multicolumn{8}{@{}l}{\textit{1,000 frames}} \\
VGGT~\cite{vggt} & 70 & 0.130 & 0.592 & 2.50 & 0.0230 & 2.26 & 373.3 \\
FastVGGT~\cite{fastvggt} & 72 & 0.100 & 0.508 & 2.06 & 0.0221 & 1.99 & 81.2 \\
LiteVGGT~\cite{litevggt}$^{\dagger}$ & 64 & 0.147 & 0.624 & 2.87 & 0.0243 & 2.68 & 43.7 \\
\rowcolor{oursrow} Eco3R (VGGT)$^{\dagger}$ & \textbf{80} & \textbf{0.093} & \textbf{0.459} & \textbf{1.43} & \textbf{0.0173} & \textbf{1.54} & \textbf{23.7} \\
  \bottomrule
\end{tabular}
}
\end{table}

\paragraph{Thousands of frames}
\Cref{stab:scale} reports the results of Eco3R at 2,000 and 4,000 frames, the two longest budgets in \cref{fig:tradeoff}. Since shorter videos retain all their frames, the 50 videos have a median length of 1,420 frames under both budgets. Three of these videos reach 4,000 frames. Under both budgets, Eco3R (VGGT) preserves its offline success rate of 82\%, while the success rate of Eco3R ($\pi^3$) remains at or above 90\%. From 2,000 to 4,000 frames, the time per frame of Eco3R remains constant. Over the same range, the trajectory error of each version of Eco3R varies by at most 0.003\,m. The peak memory of offline Eco3R (VGGT) and Eco3R ($\pi^3$) remains at 10.0 and 9.9\,GB, respectively, under both budgets (\cref{sec:method_cache}). At 2,000 frames, the full VGGT, the full $\pi^3$, FastVGGT, and FlashVGGT run out of memory in 36\%, 32\%, 60\%, and 68\% of the scenes, respectively.

\begin{table}[H]
  \caption{\textbf{Eco3R at thousands of frames on ScanNet.} Units are as in \cref{stab:scannet}. RPE is the relative rotation error, and ms/fr.\ is the runtime per frame in milliseconds.}
  \label{stab:scale}
  \centering
  \footnotesize
  \setlength{\tabcolsep}{3pt}
  \resizebox{\linewidth}{!}{\begin{tabular}{@{}l rrrrrrr@{}}
  \toprule
  Method & Frames & Succ. & ATE & RPE & Acc. & ms/fr. & Mem. \\
  \midrule
\multicolumn{8}{@{}l}{\textit{Offline}} \\
\rowcolor{oursrow} Eco3R (VGGT)$^{\dagger}$ & 2,000 & 82 & 0.101 & 0.431 & 1.54 & 46.7 & 10.0 \\
\rowcolor{oursrow} & 4,000 & 82 & 0.098 & 0.441 & 1.54 & 46.7 & 10.0 \\
\rowcolor{oursrow} Eco3R ($\pi^3$) & 2,000 & 92 & 0.060 & 0.270 & 1.03 & 40.6 & 9.9 \\
\rowcolor{oursrow} & 4,000 & 90 & 0.060 & 0.275 & 1.03 & 40.0 & 9.9 \\
\midrule
\multicolumn{8}{@{}l}{\textit{Online}} \\
\rowcolor{oursrow} Eco3R (VGGT)$^{\dagger}$ & 2,000 & 78 & 0.102 & 0.437 & 1.65 & 74.8 & 11.5 \\
\rowcolor{oursrow} & 4,000 & 78 & 0.103 & 0.435 & 1.66 & 75.1 & 11.6 \\
\rowcolor{oursrow} Eco3R ($\pi^3$) & 2,000 & 90 & 0.064 & 0.271 & 1.14 & 62.7 & 11.1 \\
\rowcolor{oursrow} & 4,000 & 92 & 0.064 & 0.272 & 1.14 & 61.8 & 11.1 \\
  \bottomrule
\end{tabular}
}
\end{table}

\paragraph{Standard ATE}
\Cref{stab:stdate} reports the standard ATE. The standard ATE is defined as the root mean square error of the camera centers after similarity alignment alone, without mapping the first predicted pose onto the first ground-truth pose. Under this metric, VGG-T$^3$, StreamVGGT, RetrieveVGGT, InfiniteVGGT, CUT3R, and TTT3R succeed in at most 58\% of the scenes. The full VGGT succeeds in 96\% and 88\% of the scenes at 500 and 1,000 frames, respectively, while offline Eco3R (VGGT) succeeds in 98\% of the scenes at both frame lengths. Offline and online, Eco3R ($\pi^3$) achieves a lower median standard ATE than every accelerated, chunked, and streaming method in \cref{stab:stdate} at both frame lengths. Eco3R (VGGT) achieves a lower median standard ATE than every method built on VGGT in \cref{stab:stdate}, both offline and online. Offline, Eco3R (VGGT) reduces the standard ATE of the full VGGT by 21\% and 28\% at 500 and 1,000 frames, respectively, with a lower error in 86\% and 82\% of the scenes. Online, Eco3R (VGGT) and Eco3R ($\pi^3$) reduce the standard ATE of CUT3R and TTT3R by 70\%--94\%, with a lower error in at least 98\% of the scenes.

\begin{table}[tb]
  \caption{\textbf{Protocol and standard ATE on ScanNet.} Succ.\ prot.\ and Succ.\ std.\ are the shares of scenes (\%) with an ATE below 0.2\,m under the protocol of FastVGGT and under the standard ATE, respectively. ATE std.\ is the standard ATE in meters. The first two rows are the full models. Shaded rows are Eco3R, while bold values denote the best results in each column of the Offline and Online groups.}
  \label{stab:stdate}
  \centering
  \footnotesize
  \setlength{\tabcolsep}{2.6pt}
  \resizebox{\linewidth}{!}{\begin{tabular}{@{}l rrr c rrr@{}}
  \toprule
  & \multicolumn{3}{c}{500 frames} & & \multicolumn{3}{c}{1,000 frames} \\
  \cmidrule(lr){2-4} \cmidrule(lr){6-8}
  Method & Succ.\ prot. & Succ.\ std. & ATE std. & & Succ.\ prot. & Succ.\ std. & ATE std. \\
  \midrule
VGGT~\cite{vggt} & 70 & 96 & 0.079 & & 70 & 88 & 0.084 \\
$\pi^3$~\cite{pi3} & 94 & 100 & 0.039 & & 94 & 100 & 0.038 \\
\midrule
\multicolumn{8}{@{}l}{\textit{Offline}} \\
FastVGGT~\cite{fastvggt} & 74 & 98 & 0.074 & & 72 & 96 & 0.079 \\
FlashVGGT~\cite{flashvggt}$^{\dagger}$ & 76 & 94 & 0.080 & & 76 & 92 & 0.081 \\
VGG-T$^3$~\cite{vggttt}$^{\dagger}$ & 14 & 30 & 0.292 & & 10 & 36 & 0.226 \\
VGGT-Long~\cite{vggtlong} & 68 & 96 & 0.083 & & 62 & 94 & 0.092 \\
\rowcolor{oursrow} Eco3R (VGGT)$^{\dagger}$ & 82 & 98 & 0.063 & & 82 & 98 & 0.061 \\
\rowcolor{oursrow} Eco3R ($\pi^3$) & \textbf{94} & \textbf{100} & \textbf{0.038} & & \textbf{94} & \textbf{100} & \textbf{0.038} \\
\midrule
\multicolumn{8}{@{}l}{\textit{Online}} \\
CUT3R~\cite{cut3r}$^{\dagger}$ & 2 & 6 & 0.582 & & 0 & 0 & 0.656 \\
TTT3R~\cite{ttt3r} & 20 & 34 & 0.230 & & 6 & 12 & 0.363 \\
StreamVGGT~\cite{streamvggt}$^{\dagger}$ & 48 & 58 & 0.166 & & \multicolumn{3}{c}{out of memory} \\
RetrieveVGGT~\cite{retrievevggt} & 38 & 50 & 0.201 & & 38 & 44 & 0.240 \\
InfiniteVGGT~\cite{infinitevggt} & 30 & 46 & 0.219 & & 20 & 36 & 0.304 \\
LASER~\cite{laser} & 84 & \textbf{100} & 0.057 & & 80 & \textbf{100} & 0.070 \\
\rowcolor{oursrow} Eco3R (VGGT)$^{\dagger}$ & 78 & 96 & 0.069 & & 78 & 96 & 0.069 \\
\rowcolor{oursrow} Eco3R ($\pi^3$) & \textbf{90} & \textbf{100} & \textbf{0.040} & & \textbf{92} & 98 & \textbf{0.042} \\
  \bottomrule
\end{tabular}
}
\end{table}

\begin{table}[tb]
  \caption{\textbf{Success rates with 95\% confidence intervals} on ScanNet (\%) for the rows of \cref{tab:main,tab:scale}. Brackets give the Wilson intervals over the 50 scenes. Rows marked failed in \cref{tab:main,tab:scale} are listed with their rates. Shaded rows are Eco3R, OOM marks out of memory, and dashes mark methods that \cref{tab:scale} omits.}
  \label{stab:successci}
  \centering
  \footnotesize
  \setlength{\tabcolsep}{3pt}
  \resizebox{\linewidth}{!}{\begin{tabular}{@{}l cccc@{}}
  \toprule
  Method & 500 frames & 1,000 frames & 2,000 frames & 4,000 frames \\
  \midrule
VGGT (full model)~\cite{vggt} & 70 [56, 81] & 70 [56, 81] & 42 [29, 56] & 42 [29, 56] \\
$\pi^3$ (full model)~\cite{pi3} & 94 [84, 98] & 94 [84, 98] & 60 [46, 72] & 60 [46, 72] \\
\midrule
\multicolumn{5}{@{}l}{\textit{Offline}} \\
FastVGGT~\cite{fastvggt} & 74 [60, 84] & 72 [58, 83] & — & — \\
FlashVGGT~\cite{flashvggt}$^{\dagger}$ & 76 [63, 86] & 76 [63, 86] & — & — \\
FlashVGGTStream~\cite{flashvggt}$^{\dagger}$ & 26 [16, 40] & 18 [10, 31] & 14 [7, 26] & 14 [7, 26] \\
VGG-T$^3$~\cite{vggttt}$^{\dagger}$ & 14 [7, 26] & 10 [4, 21] & 6 [2, 16] & 6 [2, 16] \\
Good Token Hunting~\cite{gotohunt} ($\pi^3$) & 90 [79, 96] & 92 [81, 97] & — & — \\
SAIL-Recon~\cite{sailrecon}$^{\dagger}$ & 70 [56, 81] & 72 [58, 83] & — & — \\
DA-VGGT~\cite{divpart} & 72 [58, 83] & 72 [58, 83] & — & — \\
VGGT-Long~\cite{vggtlong} & 68 [54, 79] & 62 [48, 74] & 56 [42, 69] & 56 [42, 69] \\
\rowcolor{oursrow} Eco3R (VGGT)$^{\dagger}$ & 82 [69, 90] & 82 [69, 90] & 82 [69, 90] & 82 [69, 90] \\
\rowcolor{oursrow} Eco3R ($\pi^3$) & 94 [84, 98] & 94 [84, 98] & 92 [81, 97] & 90 [79, 96] \\
\midrule
\multicolumn{5}{@{}l}{\textit{Online}} \\
CUT3R~\cite{cut3r}$^{\dagger}$ & 2 [0, 10] & 0 [0, 7] & 0 [0, 7] & 0 [0, 7] \\
TTT3R~\cite{ttt3r} & 20 [11, 33] & 6 [2, 16] & 4 [1, 13] & 4 [1, 13] \\
MUSt3R~\cite{must3r}$^{\dagger}$ & 64 [50, 76] & 68 [54, 79] & 72 [58, 83] & 72 [58, 83] \\
StreamVGGT~\cite{streamvggt}$^{\dagger}$ & 48 [35, 61] & OOM & — & — \\
RetrieveVGGT~\cite{retrievevggt} & 38 [26, 52] & 38 [26, 52] & 36 [24, 50] & 34 [22, 48] \\
InfiniteVGGT~\cite{infinitevggt} & 30 [19, 44] & 20 [11, 33] & 12 [6, 24] & 12 [6, 24] \\
LASER~\cite{laser} & 84 [71, 92] & 80 [67, 89] & 74 [60, 84] & 74 [60, 84] \\
VGGT-SLAM~\cite{vggtslam} & 50 [37, 63] & 48 [35, 61] & — & — \\
\rowcolor{oursrow} Eco3R (VGGT)$^{\dagger}$ & 78 [65, 87] & 78 [65, 87] & 78 [65, 87] & 78 [65, 87] \\
\rowcolor{oursrow} Eco3R ($\pi^3$) & 90 [79, 96] & 92 [81, 97] & 90 [79, 96] & 92 [81, 97] \\
  \bottomrule
\end{tabular}
}
\end{table}

\paragraph{Confidence intervals}
\Cref{stab:successci} lists each success rate of \cref{tab:main,tab:scale} with its 95\% Wilson interval over the 50 scenes. For paired comparisons on the same scenes, we compute the 95\% interval of the difference from 10,000 resamples of the scenes. We apply an exact McNemar test to the scenes on which only one of the two methods succeeds. At both 500 and 1,000 frames, Eco3R (VGGT) succeeds in six scenes where the full VGGT fails and in every scene where the full VGGT succeeds. The resulting gain of 12 points has a 95\% interval of 4 to 22 points (${p = 0.031}$). Relative to Eco3R (VGGT) without refocusing, the gain in success rate is 10 points (95\% interval 0 to 20 points, ${p = 0.125}$), while every error decreases significantly (\cref{stab:ablation}). Eco3R ($\pi^3$) succeeds in the same 47 scenes as the full $\pi^3$ at both frame lengths. At 500 frames, the success rates of Eco3R (VGGT-$\Omega$) and Eco3R (DA3) differ from those of their full models by 2 points, with 95\% intervals of $-4$ to 8 and $-4$ to 10 points, respectively. On VGGT and $\pi^3$, the Online Stream's success rate remains within 4 points of the offline counterpart at every frame length. The 95\% interval of each of these differences contains zero.

\section{Other datasets}
\label{ssec:heldout}

\Cref{stab:heldout} reports results on 7-Scenes, NRGBD, TUM RGB-D, and Replica. On all four datasets, we follow the 7-Scenes and NRGBD protocol of FastVGGT and subsample each sequence by a factor of two. This protocol expresses the predicted and ground-truth points in the coordinate frame of the first camera and aligns the prediction to the ground truth in scale and depth shift. The protocol then retains the central 224 $\times$ 224 pixels of each frame and registers the two point clouds by ICP before measuring the nearest-neighbor distances. We run the compared methods with the same settings as on ScanNet (\cref{ssec:methods}). On all four datasets, Eco3R ($\pi^3$) remains within 2.4\% of the full $\pi^3$ in accuracy and completeness, with no statistically significant difference in accuracy, completeness, or ATE. Eco3R ($\pi^3$) achieves speedups of 6.9$\times$, 8.6$\times$, 9.0$\times$, and 10.4$\times$ over the full $\pi^3$ on the four datasets, respectively. On every dataset, Eco3R ($\pi^3$) reduces the trajectory error of each chunked and streaming method by at least 20\%. Of these 24 differences, 20 are significant (${p < 0.05}$).

\begin{table*}[t]
  \caption{\textbf{Other datasets.} Units are centimeters for point accuracy and completeness, meters for ATE (protocol of FastVGGT), degrees for relative rotation error (RPE), and seconds per sequence for runtime. The first two rows are the full models. Shaded rows are Eco3R, while bold values denote the best results in each column of the Offline and Online groups. ``Failed'' indicates a success rate below 10\% on that dataset. $^{\ast}$Stops on 1 of 18 sequences of 7-Scenes and 2 of 12 sequences of TUM RGB-D, and the medians exclude these sequences.}
  \label{stab:heldout}
  \centering
  \footnotesize
  \setlength{\tabcolsep}{2.6pt}
  \resizebox{\textwidth}{!}{\begin{tabular}{@{}l rrrrr c rrrrr c rrrrr c rrrrr@{}}
  \toprule
  & \multicolumn{5}{c}{7-Scenes} & & \multicolumn{5}{c}{NRGBD} & & \multicolumn{5}{c}{TUM RGB-D} & & \multicolumn{5}{c}{Replica} \\
  \cmidrule(lr){2-6} \cmidrule(lr){8-12} \cmidrule(lr){14-18} \cmidrule(lr){20-24}
  Method & Acc. & Comp. & ATE & RPE & Time & & Acc. & Comp. & ATE & RPE & Time & & Acc. & Comp. & ATE & RPE & Time & & Acc. & Comp. & ATE & RPE & Time \\
  \midrule
VGGT~\cite{vggt} & 1.75 & 2.04 & 0.124 & 0.41 & 195.3 & & 1.91 & 1.30 & 0.052 & 0.22 & 136.5 & & 1.43 & 1.62 & 0.076 & 0.64 & 325.0 & & 4.70 & 2.51 & 0.182 & 0.69 & 437.8 \\
$\pi^3$~\cite{pi3} & 1.07 & 1.68 & 0.088 & 0.34 & 138.8 & & 1.27 & 0.76 & 0.030 & 0.07 & 205.7 & & 0.99 & 1.40 & 0.073 & 0.49 & 232.2 & & 1.08 & 0.76 & 0.030 & 0.10 & 313.4 \\
\midrule
\multicolumn{24}{@{}l}{\textit{Offline}} \\
FastVGGT~\cite{fastvggt} & 1.62 & 2.04 & 0.127 & 0.41 & 56.5 & & 1.78 & 1.30 & 0.045 & 0.20 & 106.0 & & 1.51 & 1.44 & 0.079 & 0.63 & 130.3 & & 3.38 & 2.18 & 0.098 & 0.42 & 134.6 \\
VGGT-Long~\cite{vggtlong} & 1.77 & 2.11 & 0.159 & 0.36 & 274.8 & & 1.82 & 1.15 & 0.075 & 0.09 & 266.9 & & 1.93 & 1.57 & 0.127 & \textbf{0.48} & 358.3 & & 1.82 & 1.40 & 0.078 & 0.12 & 344.6 \\
\rowcolor{oursrow} Eco3R (VGGT)$^{\dagger}$ & 1.18 & 2.04 & 0.100 & 0.38 & 25.3 & & 1.77 & 1.32 & 0.055 & 0.19 & 30.9 & & 1.10 & \textbf{1.37} & 0.091 & 0.62 & 32.5 & & 3.90 & 2.64 & 0.254 & 0.38 & 37.7 \\
\rowcolor{oursrow} Eco3R ($\pi^3$) & \textbf{1.08} & \textbf{1.68} & \textbf{0.088} & \textbf{0.34} & \textbf{20.1} & & \textbf{1.30} & \textbf{0.76} & \textbf{0.031} & \textbf{0.08} & \textbf{23.9} & & \textbf{1.01} & \textbf{1.37} & \textbf{0.077} & 0.56 & \textbf{25.7} & & \textbf{1.08} & \textbf{0.76} & \textbf{0.031} & \textbf{0.10} & \textbf{30.1} \\
\midrule
\multicolumn{24}{@{}l}{\textit{Online}} \\
CUT3R~\cite{cut3r}$^{\dagger}$ & failed & — & — & — & — & & failed & — & — & — & — & & failed & — & — & — & — & & failed & — & — & — & — \\
TTT3R~\cite{ttt3r} & 4.96 & 2.56 & 0.276 & 0.55 & 28.7 & & 15.50 & 7.30 & 0.416 & 0.33 & 34.9 & & 7.09 & 3.24 & 0.299 & 1.06 & 37.4 & & failed & — & — & — & — \\
InfiniteVGGT~\cite{infinitevggt} & 3.94 & 2.17 & 0.178 & 0.50 & 111.5 & & 5.47 & 1.94 & 0.172 & 0.24 & 137.6 & & 4.38 & 3.30 & 0.168 & 1.20 & 145.1 & & 9.08 & 4.41 & 0.248 & 4.18 & 171.7 \\
LASER~\cite{laser} & 2.10 & 1.70 & 0.115 & \textbf{0.33} & 46.3 & & 2.46 & 1.37 & 0.132 & \textbf{0.05} & 60.9 & & 1.87 & 2.18 & 0.128 & \textbf{0.43} & 61.6 & & 2.07 & 1.29 & 0.050 & \textbf{0.06} & 71.5 \\
VGGT-SLAM~\cite{vggtslam}$^{\ast}$ & 2.31 & 2.24 & 0.134 & 0.38 & 43.8 & & 2.35 & 1.13 & 0.114 & 0.15 & 53.1 & & 2.29 & 2.66 & 0.160 & 0.67 & 55.2 & & 5.13 & 4.43 & 0.104 & 0.14 & 67.2 \\
\rowcolor{oursrow} Eco3R (VGGT)$^{\dagger}$ & \textbf{1.20} & 1.87 & 0.098 & 0.39 & 24.8 & & 2.08 & 1.15 & 0.062 & 0.19 & 32.1 & & 1.34 & 1.59 & \textbf{0.094} & 0.70 & 35.4 & & 4.43 & 2.65 & 0.229 & 0.37 & 44.4 \\
\rowcolor{oursrow} Eco3R ($\pi^3$) & 1.29 & \textbf{1.52} & \textbf{0.092} & 0.36 & \textbf{19.2} & & \textbf{1.48} & \textbf{0.80} & \textbf{0.041} & 0.10 & \textbf{24.7} & & \textbf{1.07} & \textbf{1.46} & 0.095 & 0.66 & \textbf{27.2} & & \textbf{1.44} & \textbf{0.93} & \textbf{0.033} & 0.11 & \textbf{34.1} \\
  \bottomrule
\end{tabular}
}
\end{table*}

For Eco3R, VGGT, $\pi^3$, and FastVGGT, we measure the trajectory errors in a second run of the same configurations. The point metrics of this second run differ from those of the first run by at most 0.05\,cm in their per-dataset medians. For the online rows of Eco3R on 7-Scenes and NRGBD, we take all metrics from a single run.

\section{Online deployment and energy}
\label{ssec:deploy}

\begin{table}[H]
  \caption{\textbf{Streaming 1,000 frames} at 10 frames per second on one A100 GPU. Rate is the number of frames processed per second without waiting. Median, Max., and First are the median, maximum, and first-frame latency in seconds. Energy is the GPU energy per frame in joules. Shaded rows are Eco3R, while bold values denote the best results in each column. ``Failed'' indicates a success rate below 10\% at 1,000 frames.}
  \label{stab:online}
  \centering
  \footnotesize
  \setlength{\tabcolsep}{3.2pt}
  \begin{tabular}{@{}l rrrrr@{}}
  \toprule
  & & \multicolumn{2}{c}{Latency$\downarrow$} & & \\
  \cmidrule(lr){3-4}
  Method & Rate$\uparrow$ & Median & Max. & First$\downarrow$ & Energy$\downarrow$ \\
  \midrule
CUT3R~\cite{cut3r}$^{\dagger}$ & failed & — & — & — & — \\
TTT3R~\cite{ttt3r} & failed & — & — & — & — \\
RetrieveVGGT~\cite{retrievevggt} & 1.2 & 346.4 & 755.8 & 0.26 & 106.4 \\
InfiniteVGGT~\cite{infinitevggt} & 3.8 & 72.8 & 162.5 & 0.28 & 61.7 \\
LASER~\cite{laser} & 11.3 & 3.9 & \textbf{5.7} & 5.30 & 20.1 \\
\rowcolor{oursrow} Eco3R (VGGT)$^{\dagger}$ & 15.7 & 9.9 & 14.5 & \textbf{0.09} & 18.4 \\
\rowcolor{oursrow} Eco3R ($\pi^3$) & \textbf{18.6} & \textbf{3.2} & \textbf{5.7} & 0.22 & \textbf{15.1} \\
  \bottomrule
\end{tabular}
\end{table}

\Cref{tab:eff} lists the latency and the energy per frame at 1,000 frames. \Cref{stab:online,stab:energy} give the full measurements, taken on one 80\,GB A100 GPU as described in \cref{ssec:methods}. We stream ten of the 50 evaluation videos (every fifth one) at 10 frames per second. The full VGGT, the full $\pi^3$, VGGT-Long, StreamVGGT, RetrieveVGGT, and InfiniteVGGT run on the first five of these videos. We derive the latency of offline methods from their runtime in \cref{tab:main}. A wall clock releases frame $i$ at time $i/10$\,s. Each method can process only frames that have already been released. The latency of a frame is the time from its release until its pose and depth are available. As we decode and resize the frames beforehand, the latency excludes this cost, which all methods share.

\begin{figure}[H]
  \centering
  \includegraphics[width=\linewidth]{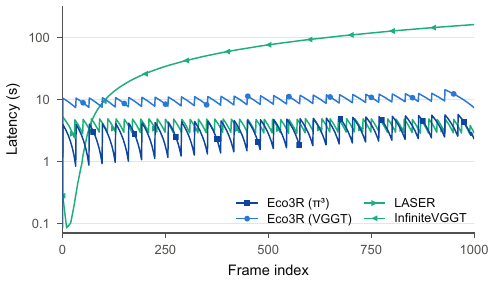}
  \caption{\textbf{Latency of every frame} when 1,000 frames arrive at 10 frames per second on one A100 GPU.}
  \label{sfig:latency}
\end{figure}

\Cref{stab:online} lists the latency at 1,000 frames and the sustainable rate, which we measure on the streamed videos. Both versions of Eco3R sustain a rate above the arrival rate of 10 frames per second. As a result, the latency of each frame remains within the Online Stream's wait of at most $(\tau+1)\kappa$ frames plus the time of the pass that emits the frame. For Eco3R ($\pi^3$) at 1,000 frames, ${\kappa = 32}$ and ${\tau = 0}$ set this wait to 32 frames, or 3.2\,s. For Eco3R (VGGT), ${\tau = 2}$ sets the wait to 96 frames, or 9.6\,s. InfiniteVGGT processes only 3.8 frames per second without waiting. In the stream, InfiniteVGGT's backlog increases by 168\,ms with every frame. StreamVGGT runs out of memory at frame 571 of 1,000. \Cref{sfig:latency} plots the median latency of every frame over the streamed videos that reach this frame (ten videos, or five for InfiniteVGGT). As each frame waits for the pass that closes its key interval and for $\tau$ further passes, the latency of Eco3R decreases from the first to the last frame of every interval of $\kappa$ frames and remains bounded.

The full models consume more energy per frame on longer videos (\cref{stab:energy}), as every frame attends to more keys. In contrast, offline Eco3R consumes the same energy per frame at both frame lengths. At 1,000 frames, offline Eco3R ($\pi^3$) consumes 11.5\,J per frame, compared with 167.0\,J for the full $\pi^3$ and 14.7\,J for FlashVGGT.

\begin{table}[tb]
  \caption{\textbf{GPU energy per frame} in joules on one A100 GPU. The first two rows are the full models. Shaded rows are Eco3R, while bold values denote the lowest energy in each column of the Offline and Online groups. ``Failed'' indicates a success rate below 10\%. Dashes mark settings without an energy measurement. $^{\ast}$Spends most of its runtime on the CPU, which this GPU measurement excludes.}
  \label{stab:energy}
  \centering
  \footnotesize
  \begin{tabular}{@{}l rr@{}}
  \toprule
  Method & 500 frames & 1,000 frames \\
  \midrule
VGGT & 110.8 & 223.4 \\
$\pi^3$ & 82.9 & 167.0 \\
\midrule
\multicolumn{3}{@{}l}{\textit{Offline}} \\
FastVGGT & 34.2 & 51.7 \\
FlashVGGT$^{\dagger}$ & 11.6 & 14.7 \\
VGG-T$^3$$^{\dagger}$ & — & 16.1 \\
VGGT-Long$^{\ast}$ & 73.7 & 72.5 \\
\rowcolor{oursrow} Eco3R (VGGT)$^{\dagger}$ & 13.7 & 13.7 \\
\rowcolor{oursrow} Eco3R ($\pi^3$) & \textbf{11.5} & \textbf{11.5} \\
\midrule
\multicolumn{3}{@{}l}{\textit{Online}} \\
CUT3R$^{\dagger}$ & failed & failed \\
TTT3R & — & failed \\
StreamVGGT$^{\dagger}$ & 255.8 & out of memory \\
RetrieveVGGT & — & 106.4 \\
InfiniteVGGT & 59.3 & 61.7 \\
LASER & \textbf{20.0} & 20.1 \\
\rowcolor{oursrow} Eco3R (VGGT)$^{\dagger}$ & 26.0 & 18.4 \\
\rowcolor{oursrow} Eco3R ($\pi^3$) & 21.2 & \textbf{15.1} \\
  \bottomrule
\end{tabular}
\end{table}

\section{Ablations, \texorpdfstring{VGGT-$\Omega$}{VGGT-Omega}, and exactness}
\label{ssec:ablation}

\begin{table}[tb]
  \caption{\textbf{Number of pinned frames} on ScanNet. Each cell gives the success rate (\%) and the median relative rotation error (degrees) of the frozen VGGT or $\pi^3$. The keys of every global block are taken from the given number of uniformly sampled frames or from all frames, as in \cref{fig:insights}a. For the frozen VGGT, the number of keys is a fixed fraction of the frames of each video. Videos that reach the frame budget use the stated numbers, with 62 in place of 63, while shorter videos use proportionally fewer keys.}
  \label{stab:keycount}
  \centering
  \footnotesize
  \setlength{\tabcolsep}{3pt}
  \resizebox{\linewidth}{!}{\begin{tabular}{@{}ll ccccc@{}}
  \toprule
  & & \multicolumn{5}{c}{Frames serving as keys} \\
  \cmidrule(lr){3-7}
  Backbone & Length & 16 & 31 & 63 & 125 & All \\
  \midrule
VGGT & 500 & 70 / 0.642 & 72 / 0.571 & 74 / 0.549 & 72 / 0.549 & 70 / 0.589 \\
VGGT & 1,000 & 72 / 0.510 & 72 / 0.501 & 72 / 0.471 & 72 / 0.454 & 70 / 0.581 \\
$\pi^3$ & 500 & 90 / 0.383 & 94 / 0.356 & 94 / 0.347 & 94 / 0.345 & 94 / 0.343 \\
$\pi^3$ & 1,000 & 92 / 0.302 & 94 / 0.281 & 94 / 0.271 & 94 / 0.267 & 94 / 0.265 \\
  \bottomrule
\end{tabular}
}
\end{table}

\paragraph{Number of pinned frames}
\Cref{stab:keycount} compares different numbers of pinned frames with full attention for the frozen VGGT and $\pi^3$ on the 50 ScanNet evaluation scenes. On the frozen VGGT at 500 frames, using 250 keys increases the relative rotation error by 2.6\% compared with using 125 keys (${p = 0.06}$), while using all 500 frames increases it by 7.2\% (${p < 10^{-10}}$). Reducing the number of pinned frames to 16 lowers the success rate of $\pi^3$ from 94\% to 90\% and 92\% at 500 and 1,000 frames, respectively. Thus, we set the number of pinned frames to ${M = 31}$, the smallest tested key set size at which the success rates of both VGGT and $\pi^3$ remain at or above those under full attention. To verify this choice, we repeated the comparison on 50 ScanNet training scans whose rooms are excluded from both our evaluation and the training of Alignment Refocusing. Across these scans, using 31 keys achieves the same success rates as using 63 and 125 keys for both backbones. With 31 keys, VGGT succeeds in 84\% and 82\% of these scans at 500 and 1,000 frames, respectively, while $\pi^3$ succeeds in 96\% at both lengths. Reducing the number of keys to 16 lowers the success rate of VGGT on these scans to 74\% at both lengths. We used the same ${M = 31}$ on the datasets of \cref{ssec:heldout}.

\begin{table}[tb]
  \caption{\textbf{Ablations of Eco3R} on ScanNet at 500 frames, in absolute values. Units follow \cref{tab:main,tab:eff}, with Comp.\ in centimeters. Gray values do not differ significantly from their reference (${p \geq 0.05}$), regardless of the size of the change. Most rows are compared with the shaded row above them, while indented rows are compared with the Online Stream row. The nearby-keys and camera-head rows are compared with their shaded configuration rerun on the same 24\,GB workstation GPU. $^{*}$Compared with a band trained with the full loss on the same 482 scans. Dashes mark costs that we do not report.}
  \label{stab:ablation}
  \centering
  \footnotesize
  \setlength{\tabcolsep}{2.4pt}
  \resizebox{\linewidth}{!}{\begin{tabular}{@{}l rrrrrr rr@{}}
  \toprule
  Variant & Succ.$\uparrow$ & ATE$\downarrow$ & RPE$\downarrow$ & Acc.$\downarrow$ & Comp.$\downarrow$ & AbsRel$\downarrow$ & TFLOPs$\downarrow$ & FPS$\uparrow$ \\
  \midrule
\rowcolor{oursrow} Eco3R (VGGT)$^{\dagger}$ & 82 & 0.090 & 0.528 & 1.48 & 1.46 & 0.0170 & 5.58 & 21.6 \\
\midrule
\multicolumn{9}{@{}l}{\textit{Components}} \\
w/o Alignment Refocusing & 72 & 0.117 & 0.571 & 2.00 & 1.76 & 0.0219 & 5.58 & 21.5 \\
w/o Pinned Key Set & 78 & 0.108 & 0.569 & 1.66 & \textcolor{gray}{1.51} & \textcolor{gray}{0.0176} & 55.59 & 2.7 \\
w/o both (VGGT) & 70 & 0.111 & 0.589 & 1.97 & 1.78 & 0.0218 & 55.59 & 2.6 \\
\midrule
\multicolumn{9}{@{}l}{\textit{Pinned Key Set}} \\
16 pinned frames & 78 & 0.103 & 0.581 & 1.63 & 1.51 & 0.0180 & 3.96 & 28.1 \\
63 pinned frames & 82 & \textcolor{gray}{0.091} & 0.510 & \textcolor{gray}{1.49} & \textcolor{gray}{1.44} & 0.0164 & 9.04 & 14.6 \\
MegaLoc keys~\cite{megaloc} & 80 & \textcolor{gray}{0.096} & \textcolor{gray}{0.529} & \textcolor{gray}{1.52} & \textcolor{gray}{1.49} & \textcolor{gray}{0.0170} & 5.69 & 21.3 \\
$+$\,own keys~\cite{kvtracker} & 80 & \textcolor{gray}{0.092} & \textcolor{gray}{0.520} & 1.62 & \textcolor{gray}{1.46} & \textcolor{gray}{0.0168} & — & — \\
$+\,8$ nearby keys & 70 & 0.124 & 0.538 & 3.12 & 2.02 & 0.0165 & 6.37 & 19.5 \\
Camera head on pinned frames & 82 & \textcolor{gray}{0.091} & \textcolor{gray}{0.536} & \textcolor{gray}{1.49} & \textcolor{gray}{1.45} & \textcolor{gray}{0.0169} & 5.58 & 21.5 \\
\midrule
\multicolumn{9}{@{}l}{\textit{Alignment Refocusing}} \\
Tuning blocks 16--21 & 80 & 0.101 & 0.576 & 1.72 & 1.52 & 0.0194 & 5.58 & 21.8 \\
Tuning blocks 0--5 & 76 & \textcolor{gray}{0.093} & 0.540 & \textcolor{gray}{1.62} & \textcolor{gray}{1.56} & \textcolor{gray}{0.0173} & 5.58 & 21.4 \\
300 training steps & 76 & 0.107 & 0.541 & 1.57 & 1.54 & 0.0181 & 5.58 & 21.6 \\
Synthetic + TUM, 300 steps & 72 & 0.114 & 0.557 & 1.76 & 1.70 & 0.0200 & 5.58 & 21.7 \\
Camera loss only$^{*}$ & 80 & 0.080 & \textcolor{gray}{0.521} & 1.65 & 1.55 & 0.0196 & 5.58 & 21.6 \\
\midrule
\multicolumn{9}{@{}l}{\textit{Online Stream}} \\
Online Stream & 78 & \textcolor{gray}{0.102} & 0.540 & 1.67 & \textcolor{gray}{1.46} & 0.0172 & 9.35 & 10.9 \\
\hspace{0.8em}translation scale & 78 & \textcolor{gray}{0.101} & \textcolor{gray}{0.540} & 1.98 & 1.62 & \textcolor{gray}{0.0172} & 9.35 & 10.7 \\
\hspace{0.8em}w/o rescaling & 66 & 0.129 & \textcolor{gray}{0.540} & 4.81 & 2.88 & \textcolor{gray}{0.0172} & 9.35 & 11.2 \\
\hspace{0.8em}translation scale, no lag & 76 & 0.101 & 0.561 & 2.38 & 1.71 & 0.0179 & 9.17 & 11.1 \\
\midrule
\rowcolor{oursrow} Eco3R ($\pi^3$) & 94 & 0.052 & 0.356 & 1.00 & 1.16 & 0.0115 & 4.79 & 25.1 \\
\midrule
\multicolumn{9}{@{}l}{\textit{Pinned Key Set}} \\
16 pinned frames & 90 & \textcolor{gray}{0.060} & 0.383 & 1.07 & 1.21 & 0.0125 & 3.55 & 31.6 \\
63 pinned frames & 94 & \textcolor{gray}{0.056} & 0.347 & 0.95 & 1.14 & 0.0110 & 7.39 & 17.7 \\
MegaLoc keys~\cite{megaloc} & 92 & 0.057 & \textcolor{gray}{0.351} & \textcolor{gray}{1.01} & \textcolor{gray}{1.18} & 0.0115 & 4.89 & 24.4 \\
$+\,8$ nearby keys & 92 & 0.073 & 0.528 & 1.43 & 1.24 & \textcolor{gray}{0.0116} & 5.38 & — \\
  \bottomrule
\end{tabular}
}
\end{table}

\paragraph{Further ablations}
\Cref{stab:ablation} lists the absolute values behind \cref{tab:ablation}, together with five additional rows. Restricting the camera head of VGGT to attend only to the pinned frames, as in the relocalization of \cref{ssec:reloc}, preserves the accuracy of Eco3R (VGGT). With this restriction, Eco3R (VGGT) succeeds in the same 82\% of the scenes as the configuration in the first row of \cref{stab:ablation}, rerun on the same 24\,GB workstation GPU, and no error changes significantly. Estimating scale from camera translation ratios and emitting frames without lag increases the accuracy error of Eco3R (VGGT) by 60.6\% relative to the offline result, while the Online Stream increases it by 12.6\% (\cref{tab:ablation,stab:ablation}). We trained the alignment band with the camera loss alone on a larger set of 482 training scans and compared it with a band trained with the full loss on the same scans. Training with the camera loss alone reduces the trajectory and relative translation errors by 17.2\% and 5.8\%, respectively, but lowers the success rate from 82\% to 80\%. The relative rotation error does not change significantly, while the accuracy error increases by 3.5\%. This suggests that the depth loss mainly contributes to the accuracy of the point clouds.

\paragraph{Layout and content of the pinned frames}
\Cref{stab:layout} compares uniformly sampled pinned frames with random and perturbed layouts of 31 frames. Sampling the pinned frames at random increases the relative rotation error of both backbones by 2.5\%--13.1\% in each of three draws (${p < 0.01}$). Averaged over the three draws of each scene, random sampling increases every error of both backbones by 4\%--21\% (${p < 0.01}$). Small departures from uniform spacing preserve pose accuracy. Jittering the pinned frames by up to a quarter of their spacing does not significantly change any trajectory or relative pose error for either backbone. Random draws place 9--16 pairs of frames less than half a spacing apart and leave gaps of up to 3.9 spacings, while jittered frames do neither. To separate these two patterns, we constructed two layouts from uniformly sampled frames, each with one pattern. The clustered layout moves 9 frames to a quarter of a spacing from a neighbor, and the layout with one long gap leaves a gap of 3.9 spacings. Either layout increases the relative translation error of both backbones by 1\%--5\% (${p < 0.001}$). Uniform sampling avoids both patterns.

Selecting the pinned frames by content performs worse than sampling them uniformly. We used farthest-point sampling on MegaLoc descriptors~\cite{megaloc} to select the 31 frames, as done in Good Token Hunting~\cite{gotohunt}, and compared the result with the uniformly sampled frames of \cref{tab:main} on the same A100 GPUs. This selection lowers the success rate of each backbone by 2 points and does not significantly reduce any error. The same selection increases the trajectory and depth errors of Eco3R ($\pi^3$) by 9.3\% and 0.6\%, respectively (${p < 0.05}$). On the frozen VGGT, selecting the pinned frames by attention or by DINOv2 descriptors~\cite{dinov2} performs worse. To select frames by attention, we measured the attention that each frame receives in block 13 during a first pass with uniformly sampled keys. Pinning the 31 most-attended frames increases the relative rotation error of the frozen VGGT from 0.57\textdegree{} to 3.75\textdegree{}. Using farthest-point sampling on DINOv2 descriptors to select the pinned frames increases the relative rotation error to 0.61\textdegree{} (${p = 0.006}$).

\paragraph{Training seeds}
We repeated Alignment Refocusing with two additional random seeds under otherwise identical settings. With each of the three seeds, Eco3R (VGGT) succeeds in 80\% or 82\% of the scenes at both 500 and 1,000 frames. The median ATE is 0.090--0.092\,m at 500 frames and 0.093--0.097\,m at 1,000 frames. Compared with Eco3R (VGGT) without refocusing, the band of each seed reduces the trajectory error by 20\%--24\% (${p \leq 0.013}$). In the Online Stream, Eco3R (VGGT) succeeds in 76\% or 78\% of the scenes at both lengths for each seed, with median ATEs of 0.098--0.102\,m. The main paper reports the first seed. To verify that the training is reproducible on other hardware, we retrained the band with the first seed on a 24\,GB workstation GPU. Evaluated on that GPU, the retrained and original bands succeed in the same 82\% of the scenes, with no significant difference in any error.

\begin{table}[tb]
  \caption{\textbf{Layout of the pinned frames} on ScanNet at 500 frames. Succ.\ is the success rate (\%), and errors are in the units of \cref{stab:scannet}. Gray values do not differ significantly from those of uniformly sampled frames (${p \geq 0.05}$). The mean-of-draws row averages each error over the three random draws of each scene. All rows ran on the same 24\,GB workstation GPU.}
  \label{stab:layout}
  \centering
  \footnotesize
  \setlength{\tabcolsep}{5pt}
  \begin{tabular}{@{}l rrrrr@{}}
  \toprule
  Layout & Succ. & ATE & RPE-r & RPE-t & Acc. \\
  \midrule
\multicolumn{6}{@{}l}{\textit{Eco3R (VGGT)}$^{\dagger}$} \\
Uniformly sampled & 82 & 0.090 & 0.528 & 1.92 & 1.48 \\
Random, draw 1 & 80 & 0.101 & 0.563 & 1.90 & 1.58 \\
Random, draw 2 & 80 & \textcolor{gray}{0.107} & 0.541 & 1.99 & \textcolor{gray}{1.52} \\
Random, draw 3 & 80 & \textcolor{gray}{0.114} & 0.578 & 2.00 & 1.60 \\
Random, mean of draws & — & 0.108 & 0.574 & 2.00 & 1.57 \\
Jittered & 80 & \textcolor{gray}{0.096} & \textcolor{gray}{0.526} & \textcolor{gray}{1.92} & 1.52 \\
One long gap & 78 & \textcolor{gray}{0.095} & \textcolor{gray}{0.534} & 1.95 & \textcolor{gray}{1.54} \\
Clustered & 80 & \textcolor{gray}{0.098} & 0.537 & 1.94 & \textcolor{gray}{1.54} \\
\midrule
\multicolumn{6}{@{}l}{\textit{Eco3R ($\pi^3$)}} \\
Uniformly sampled & 94 & 0.052 & 0.355 & 1.54 & 1.00 \\
Random, draw 1 & 92 & 0.062 & 0.377 & 1.66 & 1.06 \\
Random, draw 2 & 94 & \textcolor{gray}{0.062} & 0.369 & 1.61 & \textcolor{gray}{1.03} \\
Random, draw 3 & 94 & 0.063 & 0.402 & 1.78 & 1.10 \\
Random, mean of draws & — & 0.062 & 0.390 & 1.68 & 1.05 \\
Jittered & 92 & \textcolor{gray}{0.055} & \textcolor{gray}{0.358} & \textcolor{gray}{1.58} & \textcolor{gray}{1.00} \\
One long gap & 94 & \textcolor{gray}{0.062} & 0.370 & 1.62 & \textcolor{gray}{0.99} \\
Clustered & 94 & \textcolor{gray}{0.056} & 0.364 & 1.61 & 1.01 \\
  \bottomrule
\end{tabular}
\end{table}

\begin{table}[tb]
  \caption{\textbf{Where Alignment Refocusing acts} on ScanNet at 500 frames. Each variant reverts some tuned weights of the band to those of the frozen VGGT at inference. Succ.\ is the success rate (\%). Each error column divides the change in the median error of a variant relative to Eco3R (VGGT) without refocusing by the change achieved by Alignment Refocusing, giving the share of the gain that the variant retains (\%). Mean is the average of the four shares. Eco3R (VGGT) succeeds in 72\% of the scenes without refocusing and in 82\% with it. All runs used the same 24\,GB workstation GPU.}
  \label{stab:band}
  \centering
  \footnotesize
  \setlength{\tabcolsep}{5pt}
  \begin{tabular}{@{}l rrrrrr@{}}
  \toprule
  Variant & Succ. & ATE & RPE-r & RPE-t & Acc. & Mean \\
  \midrule
\multicolumn{7}{@{}l}{\textit{One block reverted}} \\
Block 10 & 82 & 98 & 67 & 90 & 93 & 87 \\
Block 11 & 82 & 89 & 94 & 87 & 93 & 91 \\
Block 12 & 80 & 117 & 66 & 71 & 86 & 85 \\
Block 13 & 82 & 102 & 81 & 97 & 102 & 96 \\
Block 14 & 78 & 95 & 93 & 90 & 95 & 93 \\
Block 15 & 78 & 106 & 81 & 92 & 99 & 94 \\
\midrule
\multicolumn{7}{@{}l}{\textit{Tuned blocks kept}} \\
Block 10 & 72 & 18 & 30 & 19 & 66 & 33 \\
Block 11 & 72 & 35 & 9 & 22 & 70 & 34 \\
Block 12 & 74 & 40 & 39 & 36 & 58 & 43 \\
Block 13 & 70 & 31 & $-$8 & 8 & 45 & 19 \\
Block 14 & 70 & 35 & $-$30 & 5 & 24 & 8 \\
Block 15 & 70 & 26 & 37 & 14 & 26 & 26 \\
Blocks 10--12 & 76 & 64 & 68 & 69 & 88 & 72 \\
Blocks 13--15 & 78 & 76 & 16 & 29 & 55 & 44 \\
\midrule
\multicolumn{7}{@{}l}{\textit{Tuned layers kept in all six blocks}} \\
Attention & 76 & 57 & 41 & 52 & 90 & 60 \\
MLP & 76 & 106 & 19 & 57 & 86 & 67 \\
Queries and keys & 72 & 16 & 20 & 18 & 46 & 25 \\
Values and output & 74 & 41 & 34 & 29 & 69 & 43 \\
  \bottomrule
\end{tabular}
\end{table}

\paragraph{Training with the structure of inference}
As at inference, each training sample of Alignment Refocusing contains 24 pinned frames and 24 other frames that attend only to the keys of the pinned frames (\cref{sec:method_refocus}). Let $\mathcal{P}$ denote the pinned frames of a sample, and let $\mathcal{A}$ denote its other frames. Let $\mathcal{R}$ denote six frames drawn at random from $\mathcal{P} \cup \mathcal{A}$. Alignment Refocusing minimizes
\begin{equation}
  \mathcal{L} = \lambda\, \mathcal{L}_{\mathrm{cam}}(\mathcal{P} \cup \mathcal{A}) + \mathcal{L}_{\mathrm{depth}}(\mathcal{R}),
  \label{seq:refocus}
\end{equation}
where $\mathcal{L}_{\mathrm{cam}}$ and $\mathcal{L}_{\mathrm{depth}}$ denote the camera and depth losses of VGGT~\cite{vggt}, respectively, and ${\lambda = 5}$ is the weight of the camera loss. Computing the depth loss only on $\mathcal{R}$ limits the memory required for training. We trained a second band with joint keys, in which all 48 frames attend to one another, under otherwise identical settings. Under the Pinned Key Set, both bands succeed in the same 82\% of the scenes, with no significant difference in the trajectory, relative rotation, or accuracy errors. In this setting, training with joint keys increases the relative translation error by 2.6\% (${p < 10^{-8}}$). Under full attention, the band trained with joint keys again succeeds in 82\% of the scenes, while the band trained with pinned keys succeeds in 78\%. In this setting, training with joint keys reduces the trajectory and accuracy errors by 6.4\% and 0.7\%, respectively (${p < 0.01}$). Each band achieves the lower relative translation or trajectory error under the key structure used in its training. We repeated the training of both bands on the 24\,GB workstation GPU. Under the Pinned Key Set, the two retrained bands again succeed in the same 82\% of the scenes, and training with joint keys again increases the relative translation error by 2.5\%.

\begin{table*}[b]
  \caption{\textbf{Pinned Key Set and Alignment Refocusing on other datasets.} Units follow \cref{stab:heldout}. Bold values denote the best result in each column.}
  \label{stab:heldoutabl}
  \centering
  \footnotesize
  \setlength{\tabcolsep}{2.6pt}
  \resizebox{\textwidth}{!}{\begin{tabular}{@{}l rrrr c rrrr c rrrr c rrrr@{}}
  \toprule
  & \multicolumn{4}{c}{7-Scenes} & & \multicolumn{4}{c}{NRGBD} & & \multicolumn{4}{c}{TUM RGB-D} & & \multicolumn{4}{c}{Replica} \\
  \cmidrule(lr){2-5} \cmidrule(lr){7-10} \cmidrule(lr){12-15} \cmidrule(lr){17-20}
  Method & Acc. & Comp. & ATE & RPE & & Acc. & Comp. & ATE & RPE & & Acc. & Comp. & ATE & RPE & & Acc. & Comp. & ATE & RPE \\
  \midrule
VGGT~\cite{vggt} & 1.75 & 2.04 & 0.124 & 0.41 & & 1.91 & 1.30 & 0.052 & 0.22 & & 1.43 & 1.62 & 0.076 & 0.64 & & 4.70 & 2.51 & 0.182 & 0.69 \\
\rowcolor{oursrow} Eco3R (VGGT) w/o Alignment Refocusing & 1.74 & \textbf{2.03} & 0.124 & 0.39 & & \textbf{1.34} & \textbf{1.13} & \textbf{0.030} & \textbf{0.18} & & 1.24 & 1.69 & \textbf{0.062} & 0.65 & & \textbf{2.04} & \textbf{1.62} & \textbf{0.075} & \textbf{0.31} \\
\rowcolor{oursrow} Eco3R (VGGT)$^{\dagger}$ & \textbf{1.18} & 2.04 & \textbf{0.100} & \textbf{0.38} & & 1.77 & 1.32 & 0.055 & 0.19 & & \textbf{1.10} & \textbf{1.37} & 0.091 & \textbf{0.62} & & 3.90 & 2.64 & 0.254 & 0.38 \\
  \bottomrule
\end{tabular}
}
\end{table*}

\paragraph{Where Alignment Refocusing acts}
To locate the gain of Alignment Refocusing, we reverted some tuned weights of the band to those of the frozen VGGT at inference, without retraining (\cref{stab:band}). For each variant, we measured the share of the gain that it retains, averaged over the trajectory, relative rotation, relative translation, and accuracy errors. Reverting any one of the six blocks retains 85\%--96\% of the gain, while keeping the tuned weights of any single block retains at most 43\%. Keeping only the tuned weights of blocks 10--12 or 13--15 retains 72\% and 44\% of the gain, respectively. Keeping only the tuned attention layers of all six blocks retains 60\% of the gain, while keeping only their tuned multilayer perceptron (MLP) layers retains 67\%. These results suggest that the gain of Alignment Refocusing is distributed across the whole band and across both the attention and MLP layers.

\paragraph{Pinned Key Set and Alignment Refocusing on other datasets}
\Cref{stab:heldoutabl} isolates the contributions of the Pinned Key Set and Alignment Refocusing to Eco3R (VGGT) on the datasets of \cref{ssec:heldout}. On NRGBD and Replica, where the full VGGT drifts, applying the Pinned Key Set without refocusing reduces the ATE and accuracy error of VGGT by 42\% and 30\%, respectively, on NRGBD and by 59\% and 57\% on Replica. Alignment Refocusing reduces the accuracy error of Eco3R (VGGT) by 32\% on 7-Scenes and 11\% on TUM RGB-D, with lower errors in 17 of 18 and 12 of 12 sequences, respectively. On 7-Scenes, Alignment Refocusing reduces the ATE of Eco3R (VGGT) by 20\%, with a lower ATE in all 18 sequences.

\paragraph{VGGT-$\Omega$}
VGGT-$\Omega$~\cite{vggtomega} assigns each frame 16 register tokens alongside its camera token and contains five register blocks, in which only the camera and register tokens of all frames attend to one another. The camera head of VGGT-$\Omega$ decodes the camera and register tokens of all frames jointly. We run the official code of VGGT-$\Omega$ with the checkpoint that its authors retrained for benchmarking. The official preprocessing at 416 pixels resizes the ScanNet frames to 480 $\times$ 352 pixels. Eco3R restricts the keys of the register blocks to the pinned frames, as in the global blocks. The Pinned Key Set and the Online Stream use the same settings as for Eco3R (VGGT). In single precision, the blocks that use the Scene Cache reproduce the official blocks bit for bit. On 12 frames with 4 pinned, the attached frames match a joint pass with the same pinned keys to within $2.4 \times 10^{-6}$ in pose, with a median relative depth difference of $2.4 \times 10^{-5}$. \Cref{stab:vggtomega,stab:vggtomegaother} report the results on ScanNet and on the other datasets, respectively. When running offline, Eco3R (VGGT-$\Omega$) maintains a time per frame of 24\,ms from 500 to 4,000 frames, while its success rate increases from 76\% to 82\%. On NRGBD, TUM RGB-D, and Replica, Eco3R (VGGT-$\Omega$) achieves the lowest ATE of all methods (\cref{stab:vggtomegaother,stab:heldout}). At 1,000 frames, Eco3R (VGGT-$\Omega$) requires 2.43 tera floating-point operations (TFLOPs) and 8.1\,J of GPU energy per frame offline and 3.13\,TFLOPs and 11.1\,J online, compared with 37.0\,TFLOPs and 75.6\,J for full attention. On VGGT-$\Omega$, the Online Stream sustains 28.3 frames per second and, when frames arrive at 10 per second, outputs them with a median latency of 8.9\,s.

\begin{table}[H]
  \caption{\textbf{Eco3R on VGGT-$\Omega$} on ScanNet. Shaded rows are Eco3R (VGGT-$\Omega$) offline, as in \cref{tab:ablation}. The row without the Pinned Key Set is the full VGGT-$\Omega$, which we run up to 1,000 frames, while the Online row is the Online Stream on this backbone. The time per frame (ms/fr.) is in milliseconds, and other units and notation follow \cref{tab:main,tab:scale}. Dashes mark the point clouds that we did not evaluate at 2,000 and 4,000 frames.}
  \label{stab:vggtomega}
  \centering
  \footnotesize
  \setlength{\tabcolsep}{3pt}
  \begin{tabular}{@{}l rrrrrr@{}}
  \toprule
  Variant & Succ.$\uparrow$ & ATE$\downarrow$ & RPE$\downarrow$ & Acc.$\downarrow$ & ms/fr.$\downarrow$ & Mem.$\downarrow$ \\
  \midrule
\multicolumn{7}{@{}l}{\textit{500 frames}} \\
\rowcolor{oursrow} Eco3R (VGGT-$\Omega$) & 76 & 0.112 & 0.410 & 1.95 & 23.5 & 10.9 \\
w/o Pinned Key Set & 74 & 0.107 & 0.417 & 1.97 & 128 & 30.5 \\
Online & 74 & 0.116 & 0.413 & 2.16 & 49.4 & 9.4 \\
\midrule
\multicolumn{7}{@{}l}{\textit{1,000 frames}} \\
\rowcolor{oursrow} Eco3R (VGGT-$\Omega$) & 76 & 0.106 & 0.322 & 1.98 & 23.7 & 10.9 \\
w/o Pinned Key Set & 74 & 0.109 & 0.337 & 1.98 & 249 & 55.0 \\
Online & 74 & 0.112 & 0.314 & 2.12 & 34.9 & 10.8 \\
\midrule
\multicolumn{7}{@{}l}{\textit{2,000 frames}} \\
\rowcolor{oursrow} Eco3R (VGGT-$\Omega$) & 80 & 0.104 & 0.310 & — & 24.0 & 11.0 \\
Online & 74 & 0.112 & 0.316 & — & 40.0 & 10.8 \\
\midrule
\multicolumn{7}{@{}l}{\textit{4,000 frames}} \\
\rowcolor{oursrow} Eco3R (VGGT-$\Omega$) & 82 & 0.105 & 0.315 & — & 24.1 & 13.3 \\
Online & 74 & 0.112 & 0.316 & — & 40.2 & 10.8 \\
  \bottomrule
\end{tabular}
\end{table}

\paragraph{Proof of \cref{prop:exact}}
Let $H_i^{(l)}$ denote the tokens of frame $i$ entering global block $l$. We prove by induction on $l$ that the pin and attach passes yield the same tokens as the joint pass for every frame. The initial tokens $H_i^{(0)}$ depend only on frame $i$. Only the first frame carries the reference tokens of VGGT. Since this frame is pinned, every attached frame receives the ordinary tokens, as in the joint pass. Hence, the tokens agree at block 0. Now suppose that the tokens agree at block $l$. The pin pass and every attach pass then evaluate \cref{eq:pass} on the same inputs as the joint pass. All other operations act on each frame separately. The tokens thus agree at block $l+1$. The camera head of VGGT attends across frames through one token per frame. It either runs once over all frames after the attach passes, as in the joint pass, or attends only to the pinned frames in the joint pass as well as in the pin and attach passes. Hence, all predictions agree.

We tested the exactness of \cref{prop:exact} in single precision by comparing the pin and attach passes with a joint pass whose keys are restricted to the same pinned frames. On 48 frames with 6 pinned, the poses of VGGT agree bit for bit, while the depth maps differ by at most $6.9 \times 10^{-4}$ in absolute value. This difference arises from rounding in the depth head, which convolves up to 8 frames at a time and groups the frames differently in the two computations. For $\pi^3$, the camera poses differ by at most $2.7 \times 10^{-6}$, while the local points agree bit for bit. On 8 frames with 2 pinned, the poses of DA3 differ by at most $2.4 \times 10^{-7}$, while the depths differ by a median of $4 \times 10^{-6}$ and by at most 0.13\% of the depth at individual pixels. All these differences arise from floating-point rounding. Attach batches of 32 and 128 frames reproduce the poses and depth maps of the default batches of 64 frames bit for bit, as all three sizes are multiples of 8. These two batch sizes take 1.04$\times$ and 0.99$\times$ the time of the default, respectively.

\begin{table}[tb]
  \caption{\textbf{Eco3R on VGGT-$\Omega$} on the other datasets. Point accuracy (Acc.) is in centimeters and ATE in meters, as in \cref{tab:heldout}, and the rows follow \cref{stab:vggtomega}.}
  \label{stab:vggtomegaother}
  \centering
  \footnotesize
  \setlength{\tabcolsep}{2.6pt}
  \resizebox{\linewidth}{!}{\begin{tabular}{@{}l rr c rr c rr c rr@{}}
  \toprule
  & \multicolumn{2}{c}{7-Scenes} & & \multicolumn{2}{c}{NRGBD} & & \multicolumn{2}{c}{TUM RGB-D} & & \multicolumn{2}{c}{Replica} \\
  \cmidrule(lr){2-3} \cmidrule(lr){5-6} \cmidrule(lr){8-9} \cmidrule(lr){11-12}
  Method & Acc. & ATE & & Acc. & ATE & & Acc. & ATE & & Acc. & ATE \\
  \midrule
\rowcolor{oursrow} Eco3R (VGGT-$\Omega$) & 1.96 & 0.136 & & 1.47 & 0.022 & & 1.03 & 0.051 & & 1.24 & 0.014 \\
w/o Pinned Key Set & 2.02 & 0.139 & & 1.65 & 0.023 & & 0.99 & 0.052 & & 1.38 & 0.028 \\
Online & 2.04 & 0.136 & & 1.65 & 0.022 & & 1.43 & 0.058 & & 1.30 & 0.016 \\
  \bottomrule
\end{tabular}
}
\end{table}

\section{Further analysis}
\label{ssec:analysis}

\paragraph{Query thinning}
To test whether the readout blocks can operate on fewer query tokens, we ran the frozen VGGT with all 31 pinned keys on ScanNet at 500 frames. After global blocks 16, 18, 20, and 22, we dropped nested random subsets of the patch tokens of every attached frame until half of these tokens remained. The prediction heads received the nearest retained token in place of each dropped token. Thinning the query tokens increases the median relative rotation error from 0.57\textdegree{} to 0.73\textdegree{}, with a higher error in every scene. This suggests that the readout blocks require all of their query tokens. Thus, Eco3R restricts only the keys and retains all query tokens.

\paragraph{Attention to nearby keys}
Nearby keys are frames from an attach batch that are added to the key set of that batch (\cref{sec:method_pks}). Spread keys are additional keys sampled from across the whole video. To test whether nearby keys degrade camera pose accuracy by diverting attention from the pinned keys, we measured the share of attention that each group of keys receives in the attach passes of Eco3R ($\pi^3$) on five scenes. For each group, we summed the softmax weights of its keys. With 8 nearby keys added to the 31 pinned keys, the nearby keys account for 20.5\% of the key tokens. In global blocks 13--15, the nearby keys receive 26.9\%, 30.0\%, and 28.0\% of the attention from the other attached frames, respectively, while 8 additional spread keys receive 19.7\%--20.0\%. The nearby keys receive only 6--10 percentage points more attention than their share of the key tokens but increase the relative translation error approximately five times as much as spread keys. In most other global blocks, $\pi^3$ assigns attention to each frame in proportion to its number of key tokens, to within 2 percentage points. This suggests that the attention diverted from the pinned keys accounts for only part of the degradation.

To test whether this degradation requires nearby keys to attend to the other frames of their batch, we first ran each batch without nearby keys. We then added 8 nearby keys with tokens taken from this first run, in which these frames attend only to the pinned keys. Adding these one-way keys increases the relative translation error by 63\% for Eco3R ($\pi^3$) and 85\% for Eco3R (VGGT), compared with 70\% and 80\%, respectively, for nearby keys computed in the same pass as their batch. For Eco3R ($\pi^3$), adding one-way spread keys increases the relative translation error by 33\%, compared with 13\% for spread keys computed in the same pass. This suggests that one-way keys may introduce some degradation of their own. However, whether or not nearby keys attend to the other frames of their batch, adding them increases the relative translation error far more than adding spread keys. These results suggest that the degradation is primarily attributable to the content of nearby keys, which are near-duplicate views of the same surfaces from nearby positions. The effect of one-way keys on the relative rotation error differs between the two backbones. For Eco3R ($\pi^3$), adding one-way nearby keys increases the relative rotation error by 27\%, compared with 49\% for nearby keys computed in the same pass. For Eco3R (VGGT), adding one-way nearby keys increases the relative rotation error by 17\%, compared with 2.0\% for nearby keys computed in the same pass.

\paragraph{Sharper attention}
Scaling the attention logits of every global block by 1.15 reduces the median relative rotation error of the full VGGT from 0.589\textdegree{} to 0.579\textdegree{} at 500 frames and from 0.581\textdegree{} to 0.476\textdegree{} at 1,000 frames (${p < 10^{-5}}$).

\section{Alignment Score}
\label{ssec:score}

For a single query and attention head, let $\mathbf{q}$ denote the query, and let $\mathbf{k}_1, \dots, \mathbf{k}_n$ denote the keys. Let $z_j = \mathbf{q}^{\top}\mathbf{k}_j / \sqrt{d}$ denote the logit of key $j$, where $d$ is the dimension of each attention head. Let $a_j$ denote the attention weight of key $j$. Substituting $\log a_j = z_j - \log \sum_{j'} e^{z_{j'}}$ into \cref{eq:div} gives
\begin{equation}
  \mathrm{KL}(\mathbf{u} \,\|\, \mathbf{a}) = \log \sum_{j=1}^{n} e^{z_j} - \frac{\mathbf{q}^{\top}\bar{\mathbf{k}}}{\sqrt{d}} - \log n,
  \label{seq:score}
\end{equation}
where $\bar{\mathbf{k}}$ denotes the mean of the keys. The first term is the log-sum-exp of the logits, which attention kernels already compute~\cite{flashattention}. The second term requires one dot product per query.

To select the block and the statistic of the Alignment Score, we compared two blocks (13 and 14) and two statistics of the divergence (its median and 90th percentile) on the 50 ScanNet scenes of the main paper. We chose block 13 and the 90th percentile. This setting separates the failed scenes of the full VGGT at 500 frames from the successful ones with an area under the receiver operating characteristic curve (AUROC) of 0.91. The other settings achieve AUROCs of 0.68--0.86, all higher than the AUROC of 0.61 achieved by the predicted confidence. The bootstrap 95\% confidence intervals are 0.82--0.98 for the AUROC of 0.91 and 0.14--0.48 for its margin over the AUROC of the predicted confidence.

Thresholding the Alignment Score yields a failure warning. For the full VGGT at 500 frames, the threshold that maximizes the difference between the true-positive and false-positive rates is 7.83, detecting 93\% of the failures at a false-positive rate of 17\%. In a leave-one-out test that selects the threshold for each scene on the other 49 scenes, the Alignment Score detects 87\% of the failures at the same false-positive rate. Restricting the keys increases the Alignment Score, shifting the optimal threshold from 7.83 under full attention to 9.01--9.15 for key sets ranging from a quarter to a thirty-second of the frames. As this range is narrow, we set a single threshold of 9.09 for restricted keys at 500 frames.

The Alignment Score provides information beyond room size. For the frozen VGGT with 31 pinned keys, the extent of the camera path separates failed from successful scenes with an AUROC of 0.86 when computed from the ground truth and 0.77 when estimated from the predictions. Adding the Alignment Score to a logistic regression model of failure based on the ground-truth extent significantly improves the fit (likelihood ratio test, ${p = 0.016}$).

\Cref{sec:exp_score} compares the Alignment Score with other failure signals. For the full VGGT at 500 frames, the normalized attention entropy of block 13 separates failed from successful scenes with an AUROC of only 0.52. Computed in the pin pass of the frozen VGGT with 31 pinned keys, the Alignment Score achieves an AUROC of 0.84 at a sixteenth of the cost of scoring all frames. This pin-pass score has a Spearman correlation of 0.99 with the score computed over all frames. For Eco3R (VGGT), the Alignment Score achieves an AUROC of 0.78 offline, compared with 0.55 for the predicted confidence. In the last pass of the Online Stream, the Alignment Score achieves an AUROC of 0.84.

\begin{figure*}[t]
  \centering
  \includegraphics[width=\textwidth]{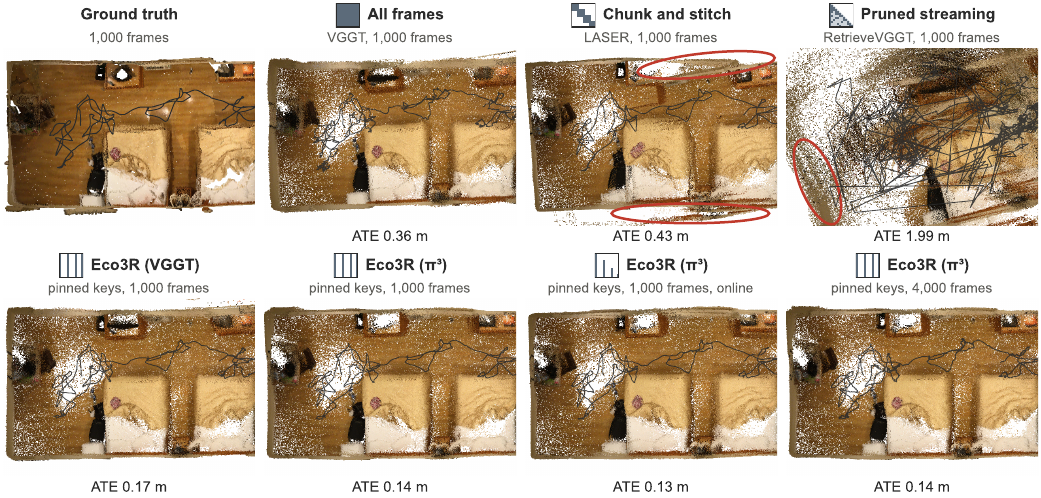}
  \caption{\textbf{Three ways to run a multi-view transformer on a long video, and Eco3R on the same video.}
  All panels show the same ScanNet video of a bedroom (scene0477\_00) from above, without the ceiling. Lines show the true camera path in the first panel and the predicted paths in the others. Icons mark which frames (columns) each frame (rows) attends to.
  Top row: with all frames as keys, every prediction lies in one coordinate frame, but the computational cost grows quadratically with the number of frames.
  Chunk and stitch bounds the cost, but errors accumulate along the chain of windows. When the camera returns to a wall, the final frames duplicate the wall (red ellipses).
  Pruned streaming reconstructs the video with bounded GPU memory as it arrives, but later frames align to the cached tokens of earlier frames, including incorrectly positioned ones. The wall that frames 100--199 incorrectly positioned remains uncorrected (red ellipse), and the room is reconstructed as rotated copies.
  Bottom row: Eco3R (VGGT) at 1,000 frames, Eco3R ($\pi^3$) offline and online at 1,000 frames, and Eco3R ($\pi^3$) offline at 4,000 frames.}
  \label{sfig:problem}
\end{figure*}

\begin{figure*}[t]
  \centering
  \includegraphics[width=0.92\textwidth]{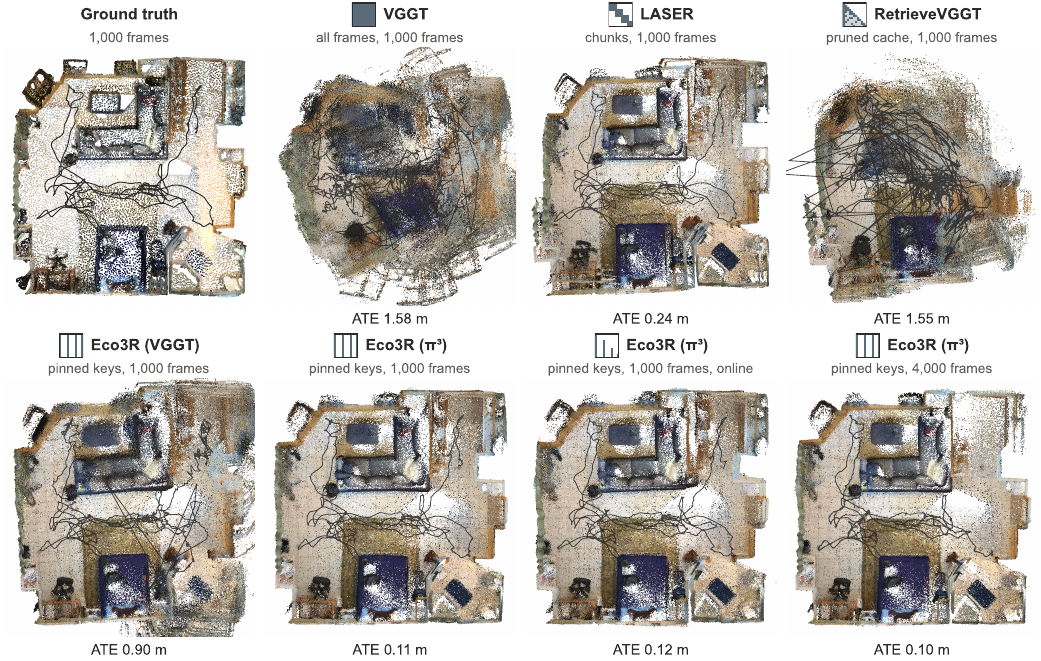}\\[6pt]
  \includegraphics[width=0.92\textwidth]{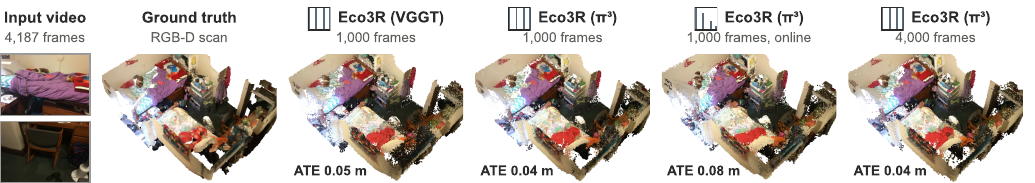}
  \caption{\textbf{More videos.} Top two rows: scene0000\_00 in top views, with the ground truth, the full VGGT, LASER, and RetrieveVGGT in the first row and the four Eco3R settings from the bottom row of \cref{sfig:problem} in the second. Lines show the camera paths, and icons show which frames each frame attends to, as in \cref{sfig:problem}. Bottom row: the top video of \cref{fig:qual} in its oblique view, with two input frames, the RGB-D ground truth, and the same four Eco3R settings.}
  \label{sfig:qual}
\end{figure*}

\section{Relocalization on 7-Scenes}
\label{ssec:reloc}

For each scene of 7-Scenes, we pin 31 frames uniformly sampled from all frames of its training sequences and attach each test image separately, with the camera head of VGGT attending only to the pinned frames. We align the pinned frames to the ground truth once per scene with a similarity transform, so that no test pose is used in this alignment. \Cref{stab:reloc} lists the median errors for each of the seven scenes, which together contain 17,000 test images. Averaged over the seven scenes, Eco3R (VGGT) requires 84\,ms per image on a 24\,GB workstation GPU and localizes 50.5\% of the test images of a scene within 5\,cm and 5\textdegree{}.

\begin{table}[H]
  \caption{\textbf{Relocalization on 7-Scenes.} Median translation (centimeters) and rotation (degrees) errors of the test images in each scene, and their mean over the seven scenes.}
  \label{stab:reloc}
  \centering
  \footnotesize
  \setlength{\tabcolsep}{3pt}
  \begin{tabular}{@{}l rrr@{}}
  \toprule
  & \multicolumn{3}{c}{Eco3R, Scene Cache of 31 frames} \\
  \cmidrule(lr){2-4}
  Scene & \begin{tabular}[b]{@{}c@{}}(VGGT) w/o\\ Alignment Refocusing\end{tabular} & (VGGT)$^{\dagger}$ & ($\pi^3$) \\
  \midrule
Chess & 3.9 / 1.83 & 3.6 / 1.75 & 3.7 / 1.83 \\
Fire & 2.9 / 1.96 & 2.8 / 1.79 & 3.5 / 2.11 \\
Heads & 1.6 / 1.92 & 1.5 / 2.09 & 1.8 / 2.07 \\
Office & 8.9 / 3.64 & 6.1 / 3.18 & 7.9 / 4.72 \\
Pumpkin & 8.1 / 3.22 & 6.9 / 2.91 & 5.7 / 2.84 \\
Red kitchen & 7.4 / 3.32 & 6.3 / 2.97 & 5.4 / 2.51 \\
Stairs & 18.2 / 2.73 & 17.1 / 2.21 & 14.2 / 2.19 \\
\midrule
Average & 7.3 / 2.66 & 6.3 / 2.41 & 6.0 / 2.61 \\
  \bottomrule
\end{tabular}
\end{table}

\section{More videos}
\label{ssec:videos}

\Cref{sfig:problem} shows how three common ways of running a multi-view transformer fail at 1,000 frames on scene0477\_00 (the bottom video of \cref{fig:qual}) and how Eco3R reconstructs the same video. LASER and RetrieveVGGT duplicate walls, while Eco3R reconstructs each wall exactly once. \Cref{sfig:qual} shows Eco3R on scene0000\_00 and on the top video of \cref{fig:qual}. On scene0000\_00, the full VGGT has an ATE of 1.58\,m, while Eco3R ($\pi^3$) achieves ATEs of 0.10--0.12\,m offline and online at 1,000 frames and offline at 4,000 frames. On the top video of \cref{fig:qual}, Eco3R (VGGT) and the same three settings of Eco3R ($\pi^3$) achieve ATEs of 0.04--0.08\,m.

\section{Limitations and future work}
\label{ssec:limits}

\paragraph{Remaining failures}
The remaining failures of Eco3R (VGGT) occur in large rooms (\cref{sec:method_refocus}), such as scene0000\_00 in the second row of \cref{sfig:qual}. None of the four modifications to the network and its inputs that we tested reduced these failures. We first added a bias to the attention logits between pinned frames in the alignment band. This bias equals the co-visibility of the two frames multiplied by a coefficient $\beta$. We computed the co-visibility from either the ground truth or the predicted geometry. Across the 50 scenes, setting ${\beta = 1}$ increases the trajectory error of the pinned frames by at most 3.9\%, while setting ${\beta = 4}$ increases it by 33\%--94\%. Learning this bias during Alignment Refocusing, with one $\beta$ per block and head initialized to zero, raises the mean of the learned $\beta$ only to 0.0035, while the success rate remains at 82\%. Passing the pinned frames through the alignment band two or three times, with learned gates initialized to zero, leaves the gates near 0.01 after training and the success rate at 82\%. Widening the inputs beyond the 518 pixels used in training increases the trajectory error of both versions of Eco3R. At 500 frames, using inputs 644 and 700 pixels wide lowers the success rate of Eco3R (VGGT) from 82\% to 76\% and 70\% and increases its trajectory error by 45\% and 56\%, respectively. At 700 pixels, Eco3R ($\pi^3$) preserves its success rate of 94\%, but its trajectory error increases by 23\%.

The remaining errors lie mainly in the poses of the pinned frames, to which the attached frames are aligned. Replacing only the poses of the 31 pinned frames with the ground truth and moving each attached frame by the correction interpolated from its neighboring pinned frames removes 8 of the 9 failures at 500 frames. Neither adding loop edges from two-frame passes of the network nor using mirrored inputs removes these failures, as the resulting pose errors point in nearly the same directions as those of the pin pass. Applying bundle adjustment over feature tracks increased the errors in the two scenes that we tested, as the tracks contain mismatches across the wide baselines between pinned frames. Future work will constrain the pinned poses with wide-baseline matches whose errors are not correlated with those of the network.

\begin{table}[tb]
  \caption{\textbf{KITTI odometry.} ATE (meters) after alignment by a similarity transform alone. Under Eco3R ($\pi^3$), Online denotes the Online Stream with ${\kappa = 16}$ and ${M_{\max} = 32}$, Recent keys denotes the same stream with keys from the 32 most recent candidates, and Offline denotes the Pinned Key Set with 31 keys.}
  \label{stab:kitti}
  \centering
  \footnotesize
  \setlength{\tabcolsep}{3pt}
  \resizebox{\linewidth}{!}{\begin{tabular}{@{}l r rrr r@{}}
  \toprule
  & & \multicolumn{3}{c}{Eco3R ($\pi^3$)} & \\
  \cmidrule(lr){3-5}
  Sequence & Frames & Online & Recent keys & Offline & LongStream~\cite{longstream}$^{\dagger}$ \\
  \midrule
00 & 4,541 & 179.7 & 177.2 & 189.6 & 89.3 \\
01 & 1,101 & 405.9 & 256.7 & 282.9 & 63.6 \\
02 & 4,661 & 301.8 & 265.8 & 301.5 & 236.9 \\
03 & 801 & 41.6 & 21.6 & 6.7 & 5.2 \\
04 & 271 & 6.9 & 6.9 & 2.3 & 2.5 \\
05 & 2,761 & 152.1 & 50.4 & 156.4 & 80.9 \\
06 & 1,101 & 38.9 & 30.9 & 28.9 & 12.8 \\
07 & 1,101 & 50.5 & 45.4 & 47.7 & 13.9 \\
08 & 4,071 & 240.5 & 121.8 & 251.9 & 89.5 \\
09 & 1,591 & 209.8 & 115.0 & 213.4 & 95.5 \\
10 & 1,201 & 175.5 & 69.7 & 206.7 & 25.7 \\
\midrule
Mean & & 163.9 & 105.6 & 153.4 & 65.1 \\
Median & & 175.5 & 69.7 & 189.6 & 63.6 \\
  \bottomrule
\end{tabular}
}
\end{table}

\paragraph{Outdoor driving}
\label{ssec:kitti}
We run Eco3R ($\pi^3$) online with ${\kappa = 16}$ and ${M_{\max} = 32}$ on the 11 sequences of KITTI odometry that have ground truth. We align each trajectory to the ground truth with a similarity transform and list the errors in \cref{stab:kitti}. As a car rarely revisits a place, most keys share no content with the current frame. Running Eco3R ($\pi^3$) online yields a mean ATE of 163.9\,m. Using keys from the 32 most recent candidates reduces the mean ATE to 105.6\,m, while running Eco3R ($\pi^3$) offline yields a mean ATE of 153.4\,m. LongStream~\cite{longstream} achieves a mean ATE of 65.1\,m in our run of its official code. With recent keys, the Online Stream of Eco3R ($\pi^3$) requires a median of 33\,ms per frame and at most 9.7\,GB of peak memory. Future work will select keys by co-visibility and add loop closure.

\paragraph{Alignment Score}
With 31 keys, the threshold of the Alignment Score depends on the video length. For the frozen VGGT with 31 keys at 1,000 frames, the threshold of 9.09 detects 57\% of the failures at a false-positive rate of 11\%. With 125 and 62 keys, the same threshold detects 93\% and 71\% of the failures at false-positive rates of 17\% and 11\%, respectively. Future work will calibrate the threshold jointly over video lengths and key counts. For $\pi^3$, we measure the divergence in block 14, as the divergence of both backbones peaks around this block (\cref{sec:method_band}). In this block, the attention of $\pi^3$ remains sharp in every scene. The 90th percentile of the divergence is 33.5 in failed scenes and 33.1 in successful scenes, leaving the Alignment Score unable to detect the failures of $\pi^3$. Future work will search for a failure signal of $\pi^3$ in its other blocks and its camera head.

\paragraph{Synthetic scenes}
Alignment Refocusing increases the accuracy and trajectory errors of Eco3R (VGGT) on NRGBD and Replica (\cref{stab:heldoutabl}). We train the alignment band only on ScanNet, whose field of view is approximately 58\textdegree{}. On Replica, whose field of view is 73.7\textdegree{}, refocusing narrows the predicted field of view by 4.5\textdegree{}. This suggests that refocusing may bias the predicted field of view toward that of the training data. The gain of refocusing on ScanNet likewise depends on the training domain. After 300 steps, training on ScanNet scans improves the success rate of Eco3R (VGGT) to 76\%, while training on synthetic scenes and TUM RGB-D leaves the success rate at the 72\% achieved without refocusing. Future work will train the alignment band on diverse cameras and scenes.

\paragraph{Online results}
Running online increases the accuracy error of Eco3R (VGGT) and Eco3R ($\pi^3$) by 7\%--22\% on ScanNet and by 2\%--33\% on the four other datasets, relative to the offline results. On TUM RGB-D, running online increases the standard ATE of Eco3R (VGGT) and Eco3R ($\pi^3$) by 39\% and 36\%, respectively, with a higher error in 11 of the 12 sequences for both versions. On the other three datasets, the standard ATE increases by 4\%--13\%. As each pass normalizes the scale according to its own frames, the online result depends on how the passes are stitched. For example, estimating scale from camera translation ratios increases the accuracy error by 18.6\% compared with depth-median-based estimation. Omitting rescaling altogether increases the accuracy error by 189\% (\cref{tab:ablation}). Future work will refine the scales of emitted passes as more of the video arrives.

\paragraph{Relative poses}
For $\pi^3$, the Pinned Key Set increases the relative pose errors on ScanNet by 4\%--11\% and the relative rotation error on the other datasets by up to 14\%, as using 31 keys provides each frame with fewer views than using all frames. The Pinned Key Set increases the standard ATE of $\pi^3$ in 38 and 35 of the 50 ScanNet scenes at 500 and 1,000 frames, respectively. With 125 keys, the gap to the full $\pi^3$ model in the relative pose errors on ScanNet narrows to at most 1.2\% (\cref{stab:keycount} lists the relative rotation errors). Future work will adapt the number of pinned frames to each video and add pinned frames where neighboring pinned frames share little content.

\FloatBarrier

\end{document}